\documentclass[11pt,a4paper]{article}
\usepackage[margin=22mm]{geometry}
\usepackage[T1]{fontenc}
\usepackage[utf8]{inputenc}
\usepackage{lmodern,microtype}
\usepackage{comment}
\usepackage{amsmath,amssymb,mathtools}
\usepackage{graphicx,booktabs,tabularx,array}
\usepackage[round,authoryear]{natbib}
\usepackage[font=small,labelfont=bf]{caption}
\usepackage{xcolor,enumitem,placeins}
\usepackage[hidelinks]{hyperref}
\hypersetup{pdftitle={Beyond the record:  evaluating probabilistic AI weather forecasts deep in the tail},pdfkeywords={AI weather, records, CRPS, probabilistic verification, reference sensitivity}}
\newcolumntype{P}[1]{>{\raggedright\arraybackslash}p{#1}}
\newcolumntype{Y}{>{\raggedright\arraybackslash}X}
\newcommand{\ind}{\mathbf{1}}
\newcommand{\E}{\mathbb{E}}

\newcommand{\CRPS}{\operatorname{CRPS}}
\newcommand{\rtw}{\operatorname{rtwCRPS}}
\newcommand{\fCRPS}{\widehat{\operatorname{CRPS}}}

\title{\textbf{Beyond the record: evaluating probabilistic AI weather forecasts deep in the tail}}

\author{Bastiaan Van Velthoven, Cas Decancq, Diego G. Miralles, Stijn Luca, Thomas Mortier}

\date{}

\begin{document}
	\maketitle
	\vspace{-12mm}
	\begin{abstract} 
		Can probabilistic artificial intelligence (AI) weather forecasts retain skill beyond their training domain? We compare four state-of-the-art AI models (ArchesWeatherGen, U-Cast, WeatherNext2 and GenCast-mini) with the physics-based IFS-ENS over land in 2022, at 1.5$^\circ$ resolution and lead times of 2--10 days. Monthly ERA5 records from 1979--2021 define common historical boundaries, and we characterize new records set in 2022 by their depth, i.e., the exceedance margin relative to local historical variability. Two proper scores, record-threshold weighted CRPS and Brier score, assess the magnitude and probability of record exceedances across all eligible cases, retaining both missed extremes and false alarms. Verified against ERA5, AI models outperform IFS-ENS for heat and cold records at almost every lead and depth. For heat, the advantage strengthens with depth, partly because IFS-ENS over-forecasts deep records, and largely persists for deeper records under HRES-fc0 verification. For wind, gains erode for deeper records and longer leads. Better tail scores can nevertheless coexist with aggregate probability deficits or surpluses, and some improvements arise chiefly on non-exceedances. Revisiting deterministic forecasts for 2020, we find that model rankings can reverse when only the verification reference changes, so that the choice of reference is itself part of the result. Increasing underprediction on progressively more extreme observed subsets can also arise through outcome selection, even for an ideal conditional-mean forecast. These findings establish that native probabilistic AI ensembles retain, and for heat increase, their advantage over IFS-ENS as thresholds move beyond local records, although all advantages narrow with lead time. They caution against generalizing the extrapolation failure reported for deterministic AI forecasts to probabilistic systems, and motivate common verification conventions that separate comparative tail skill, performance on realized extremes, and probabilistic calibration.
	\end{abstract}
	
    \section{Significance statement} 
    
Adopting AI for weather forecasting requires evidence that these systems can anticipate unprecedented conditions, an increasingly urgent question in a warming climate. Earlier studies of deterministic forecasts raised concerns about predicting record-breaking extremes. We show that probabilistic systems, which forecast a range of possible outcomes, outperform even leading physics-based systems across most variables and lead times examined, even beyond local historical records. These findings challenge claims that AI generally fails beyond historical extremes. We also show that the verdict on a forecast system can reverse depending on which dataset is taken as the truth, and call for verification practices and open forecast archives that make such comparisons reproducible.
    
	\section{Introduction} 

    Machine learning models are expected to struggle beyond the range of their training data: relationships learned from finite samples need not hold under conditions that were never observed \citep{liu2023ood}. Extrapolating skillfully requires explicit assumptions about how those relationships continue past the data \citep{buritica2024progression,pfister2026extrapolation,shen2025engression,engelke2026evt}. This concern has become urgent in weather forecasting. In only a few years, AI weather models have come to rival or outperform leading physics-based systems such as the European Centre for Medium-Range Weather Forecasts' (ECMWF) Integrated Forecasting System (IFS) on standard benchmarks \citep{pathak2022fourcastnet,bi2023pangu,lam2023graphcast,chen2023fuxi,couairon2024archesweather,lang2024aifs,bodnar2025aurora,rasp2024weatherbench2}, prompting a ``second revolution in weather forecasting'' \citep{pasche2025validating}. Unlike physics-based models, which solve equations of atmospheric motion that have no special boundary at a historical record, AI models learn from a few decades of past states. Whether their skill carries over to weather that has never been seen before therefore remains an open question \citep{magnusson2023_ml_extremes_ecmwf,ben2024rise,pasche2025validating}.

    The question matters most for record-breaking extremes, because these are where forecasts are needed most. Extreme heat is among the deadliest climate hazards \citep{zhao2021global,ballester2023heat,EEA2025_ClimateStateOutlook_3_2}, and cold spells and windstorms likewise strain infrastructure and preparedness measures designed around past experience. Communities that have rarely faced severe extremes are especially vulnerable when long-standing records fall \citep{fischer2021shattering,thompson2023atrisk}. Climate change is making such events more common: hot records over land are now broken several times more often than expected in a stable climate, and precipitation records are increasing as well \citep{fischer2025review}. Records are projected to be broken by increasingly large margins \citep{fischer2021shattering}, and unprecedented heat is emerging fastest in tropical regions where natural variability is small \citep{coumou2013records,robinson2021extremes}. Early warnings can reduce loss of life and economic damage \citep{WMO2015Heatwaves,vsakic2022early,WMO2025_COP30_ClimateUpdate}, but only if forecasts remain skillful precisely when conditions move beyond historical experience. 

    The evidence so far is discouraging. \citet{zhang2026recordextremes} conducted a first global assessment of AI weather forecasts for record-breaking extremes. They defined local records of hot, cold and windy conditions in ERA5 \citep{hersbach2020era5} during 1979--2017, and evaluated how well GraphCast, Pangu-Weather and FuXi forecast subsequent record exceedances, compared with ECMWF's high-resolution forecast (HRES; \citealp{ecmwf2021ifs}). Despite their strong overall performance, the AI models underestimated both how often records were broken and by how much, with errors growing as records were exceeded by larger margins. The authors interpreted this as evidence that AI weather models are structurally limited in their ability to extrapolate to unprecedented conditions. 
    
    We argue that these results do not by themselves support such a strong conclusion. The models evaluated were deterministic, and therefore do not learn the full predictive distribution. Because the atmosphere is chaotic, small uncertainties in the initial state grow into a range of plausible futures \citep{Lorenz1969,judt2018insights,selz2022transition}. A model trained to minimize squared error learns to predict the average of these futures \citep{gneiting2011point}, which smooths the forecast and dampens local peaks \citep{selz2026resolution,decancq2026weather,subich2025fixing,brenowitz2025benchmark,ebert2013progress}. Moreover, when evaluation is restricted to events that turned out to be extreme, even a perfect forecast of the average will systematically underestimate them \citep{lerch2017forecasters}. Underprediction that grows with the record margin is therefore expected, regardless of whether a model can extrapolate to unprecedented conditions (Supplementary Section~\ref{sec:detconditional}). A fairer test asks whether a forecast assigns meaningful probability to a record-breaking outcome, even when its average stays within the historical range. There are also good reasons to expect that AI models can pass this test. \citet{biegert2026weighted} showed that, once each forecast is statistically recalibrated in the same way \citep{gneiting2026potential}, deterministic AI forecasts often contain more information about extreme outcomes than HRES.  
    
    Recently, a new generation of AI ensemble systems has been proposed that learn the full predictive distribution by means of diffusion and flow-matching methods \citep{price2025gencast,couairon2026arches,chen2025fengwu}, training on probabilistic scores \citep{alet2025skillful,zhong2025fuxiens,ruhlingcachay2026ucast,lang2026aifscrps}, or combining learned components with atmospheric dynamics \citep{kochkov2024neuralgcm}. Compared to earlier efforts, these systems give rise to sharp, physically plausible ensemble forecasts rather than a single average. Several already rival or outperform the physics-based IFS ensemble (IFS-ENS) on standard probabilistic benchmarks \citep{price2025gencast,zhong2025fuxiens,bonev2025fourcastnet3,lang2026aifscrps,rasp2026weathernext}. Whether these ensembles retain their advantage for record-breaking events, without any additional recalibration, has not yet been tested. 

\begin{figure}\centering\includegraphics[width=0.7\linewidth]{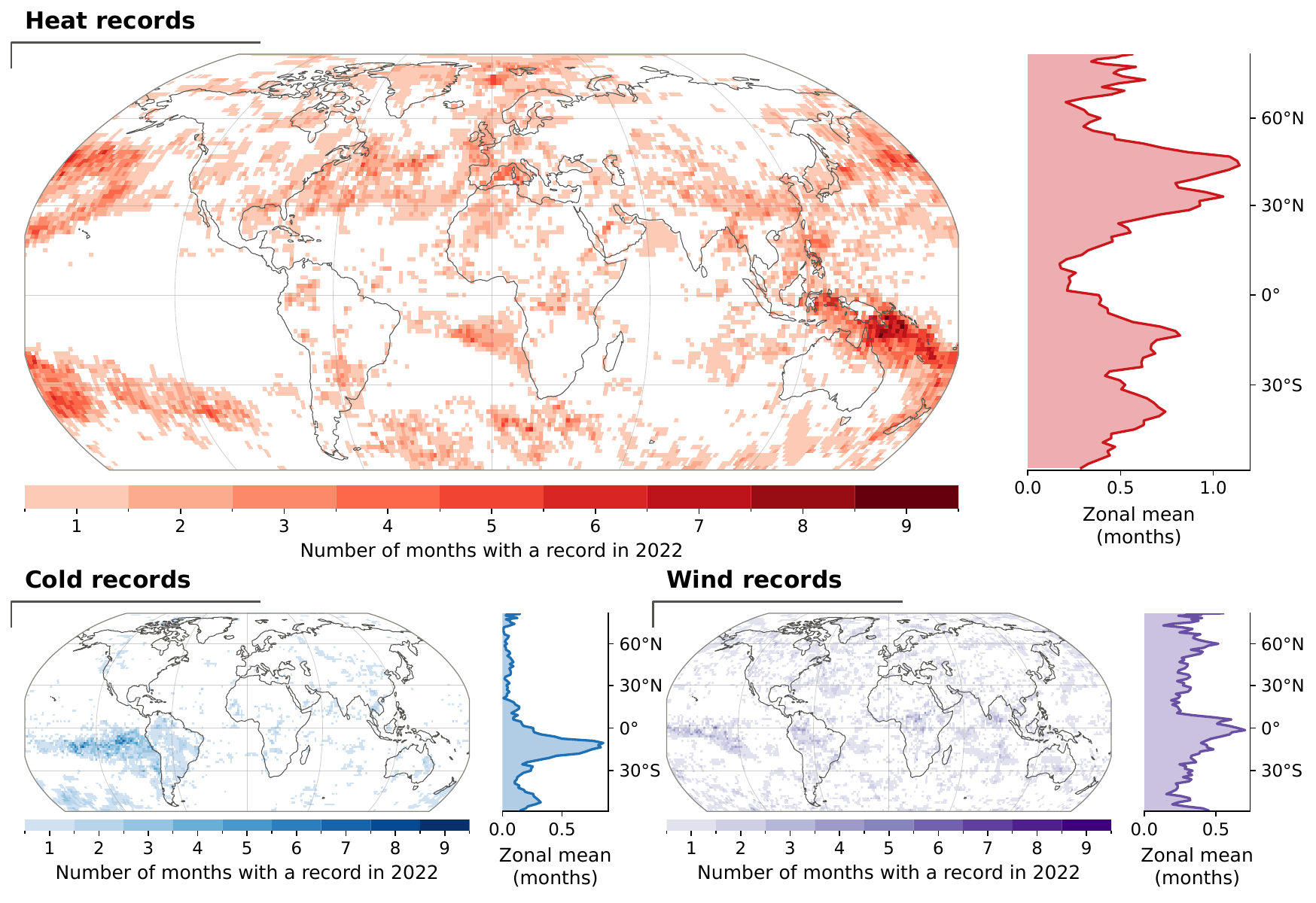}
        \caption{\textbf{New ERA5 records set in 2022.} For heat (top), cold (lower left), and wind speed (lower right), the figure shows how many months of the year in 2022 experienced an extreme that went beyond historical (1979-2021) monthly ERA5 records.}
		\label{fig:records2022}
    \end{figure}

    Here we address this question by benchmarking four AI ensemble systems against IFS-ENS: ArchesWeatherGen (AWG; \citealt{couairon2024archesweather}), U-Cast \citep{ruhlingcachay2026ucast}, WeatherNext~2 (WN2; \citealt{rasp2026weathernext}), and GenCast-mini \citep{price2025gencast}. We evaluate forecasts of temperature and wind over land in 2022, at lead times from two to ten days. Local monthly records are defined from ERA5 during 1979--2021, and evaluation thresholds are then pushed progressively further to subset new records set in 2022 (Figure~\ref{fig:records2022}), in steps scaled by local historical variability. At each threshold, we score both the predicted probability of exceeding it and the predicted severity of exceedances, by means of the Brier score and threshold-weighted CRPS score, respectively~\citep{gneiting2011comparing}. We then examine where differences in skill arise, by separating the contributions of events that did and did not break the threshold, and test how sensitive the comparison is to the reference used for verification (ERA5 versus HRES analyses). Finally, we revisit deterministic forecasts for 2020 to disentangle the effects of event selection from those of the verification reference. Together, these analyses test whether the extrapolation failure reported for deterministic AI forecasts \citep{zhang2026recordextremes} carries over to the AI ensemble systems now entering operational use.

	\section{Data} 
    
	\subsection{Verification data}
	\label{sec:groundtruth}
	
	ERA5 is our primary verification reference and is used to create historical record thresholds and other statistics related to local variability. It reconstructs the global atmosphere by combining observations with a consistent modelling and data assimilation system \citep{hersbach2020era5}. Its 1979--2021 historical boundaries form the base to identify new records set in 2022 against (Figure~\ref{fig:records2022}, Section~\ref{sec:supp_records}). We also use HRES-fc0 from WeatherBench~2 \citep{rasp2024weatherbench2}. These are the regridded zero-hour forecast fields from the high-resolution configuration of ECMWF's Integrated Forecasting System \citep{ecmwf2021ifs}, although no comparable historical archive is available. As two distinct approximations of reality, these two datasets do not share an exact climatology (Section~\ref{sec:supp_records}). Neither product is an error-free observation of the atmosphere. Verification can be sensitive to such errors in the reference data and to dependence between forecasts and their verifying analyses \citep{ferro2017measuring,bessac2021forecast,ramon2024uncertainties, caron2025reliability}.
	
	We evaluate 2~m temperature and 10~m wind speed in 2022 on a common 1.5$^\circ$ grid. Heat and cold refer to upper and lower temperature extremes, respectively, while wind extremes concern high wind speeds. Following \citet{zhang2026recordextremes}, we restrict verification to land and exclude Antarctica, retaining cells with ERA5 land fraction greater than 0.5 and latitude greater than $-60^\circ$. The available 2022 WN2 archive does not include the December 2021 initializations needed for 240-hour forecasts verifying during the first ten days of January (Section~\ref{sec:probforecasts}). Evaluating twice daily at 00 and 12~UTC, we retain 710 valid times from January 11$^{th}$ to December 31$^{st}$ 2022.
	
	Following \citet{gneiting2026potential} and \citet{biegert2026weighted}, we evaluate all models against a common reference and repeat the comparison using both ERA5 and HRES-fc0, where historical records and local variability are always derived from ERA5. Only the verifying values and corresponding exceedance labels change; HRES-fc0 exceedances therefore represent crossings of ERA5 historical thresholds (Section~\ref{sec:supp_records}). Unless stated otherwise, results use ERA5. We return to the implications of reference sensitivity in the Discussion.
	
	\subsection{Probabilistic forecasts}
	\label{sec:probforecasts}
	
	We compare four AI ensemble systems, 
	AWG, U-Cast, WN2, and GenCast-mini, with ECMWF's Integrated Forecasting System ensemble, IFS-ENS. Model selection is guided by access to ensemble forecast archives or public model checkpoints. These resources remain less widely available than for deterministic forecasts, particularly for a common evaluation year beyond all model training periods. We obtain IFS-ENS and WN2 forecasts for 2022 from WeatherBench~2. We generate AWG, U-Cast, and GenCast-mini forecasts using the authors' open-source code \citep{couairon2025geoarches, cachay2026ucast_code, gencastmini2026}. We evaluate lead times of 48, 120, 168 and 240~h, corresponding to 2, 5, 7 and 10 days, using 64 members for AWG, U-Cast and WN2, 32 for GenCast-mini, and 50 for IFS-ENS.
	
	AWG combines a transformer-based deterministic forecaster with residual flow matching at 1.5$^\circ$ resolution \citep{couairon2026arches}. Its deterministic component is trained with mean squared error on ERA5 data through 2018, while the generative stage includes 2019. U-Cast uses a standard U-Net backbone at the same resolution, trained on ERA5 data from 1979 to 2019 by deterministic pre-training on mean absolute error, followed by short probabilistic fine-tuning on the fair CRPS, with Monte Carlo dropout and deep ensembles providing the ensemble members \citep{ruhlingcachay2026ucast}. WN2 generates ensembles through learned functional perturbations and deep ensembles, using the fair CRPS objective \citep{alet2025skillful}. We use the operational forecasts available from WeatherBench~2 at 1.5$^\circ$ resolution. This configuration is pretrained on ERA5, then fine-tuned and initialized using HRES-fc0 \citep{wb2faq2026}. The checkpoint underlying the 2022 archive was trained through 2021.
	
	We include GenCast-mini as a deliberately lightweight configuration of the diffusion-based GenCast forecasting system \citep{price2025gencast,gencastmini2026}. It uses a coarser 1$^\circ$ grid and a reduced internal mesh compared with full-resolution (0.25$^\circ$) GenCast; here, we evaluate a 32-member ensemble. Its checkpoint is trained on ERA5 during 1979--2018 and initialized from ERA5, giving it the earliest final training cutoff among the AI systems considered. Forecasts are regridded to the common 1.5$^\circ$ evaluation grid following \cite{rasp2024weatherbench2}. Its inclusion broadens the benchmark to a computationally accessible diffusion model evaluated several years after training ends, but its results should not be generalized to the larger GenCast configurations.
	
	The different training histories require care when interpreting model rankings. WN2 was trained on more recent data and uses a different initialization reference, so any advantage cannot be attributed uniquely to its modelling approach. The common 1979--2021 ERA5 record baseline gives every system the same historical thresholds, but does not place them at an equal distance beyond their respective training extremes. We therefore assess the available forecasting systems as a whole, including their training histories, initialization sources and model designs.
	
	\section{Methods} 
    \subsection{Historical records, margins and record depth}

Let $x_{g,t}$ denote the variable of interest, such as 2-m temperature, at grid cell $g$ and time $t$. We define historical records using ERA5 data at 00 and 12~UTC during 1979--2021, denoting this set of times by $T_{\mathrm{hist}}$. For each grid cell $g$ and calendar month $m$, we define

\begin{equation}
	r^{\max}_{g,m}
	= \max_{\substack{t\in T_{\mathrm{hist}}\\m(t)=m}} x_{g,t},
	\qquad
	r^{\min}_{g,m}
	= \min_{\substack{t\in T_{\mathrm{hist}}\\m(t)=m}} x_{g,t},
	\label{eq:records}
\end{equation}

where $m(t)$ denotes the calendar month of $t$. The corresponding variability scales, $s^{\max}_{g,m}$ and $s^{\min}_{g,m}$, are the sample standard deviations across the 43 historical years of the monthly maxima and minima, respectively. Hereafter, for ease of notation, $i$ indexes verification cases, each associated with a grid cell $g_i$ and verification time $t_i$. 

We express record margins in units of local variability to account for differences in climate variability associated with latitude, elevation, land cover, and other factors. The scaled record depth $d_i$ and corresponding thresholds $u_i(k)$ are

\begin{equation}
	d_i=\pm\frac{x_i-r_i}{s_i},
	\qquad
	u_i(k)=r_i\pm k s_i,\quad k\geq 0,
	\label{eq:depth}
\end{equation}

where both expressions use $+$ for upper extremes and $-$ for lower extremes. Thus, $d_i>0$ indicates a record-breaking observation, while $d_i>k$ identifies observations more than $k$ local standard deviations beyond the historical record. Increasing $k$ therefore targets increasingly rare events farther beyond the historical record.

	\subsection{Proper scores for ensemble forecasts}
	
	For a predictive cumulative distribution function $F$ and verifying observation $x$, the continuous ranked probability score is
    
	\begin{equation}
		\CRPS(F,x)
		=\int_{-\infty}^{\infty}
		[F(z)-\ind\{x\le z\}]^2\,\mathrm dz.
	\end{equation}
    
\begin{figure}[htbp]
    \centering
    \includegraphics[width=\linewidth]
    {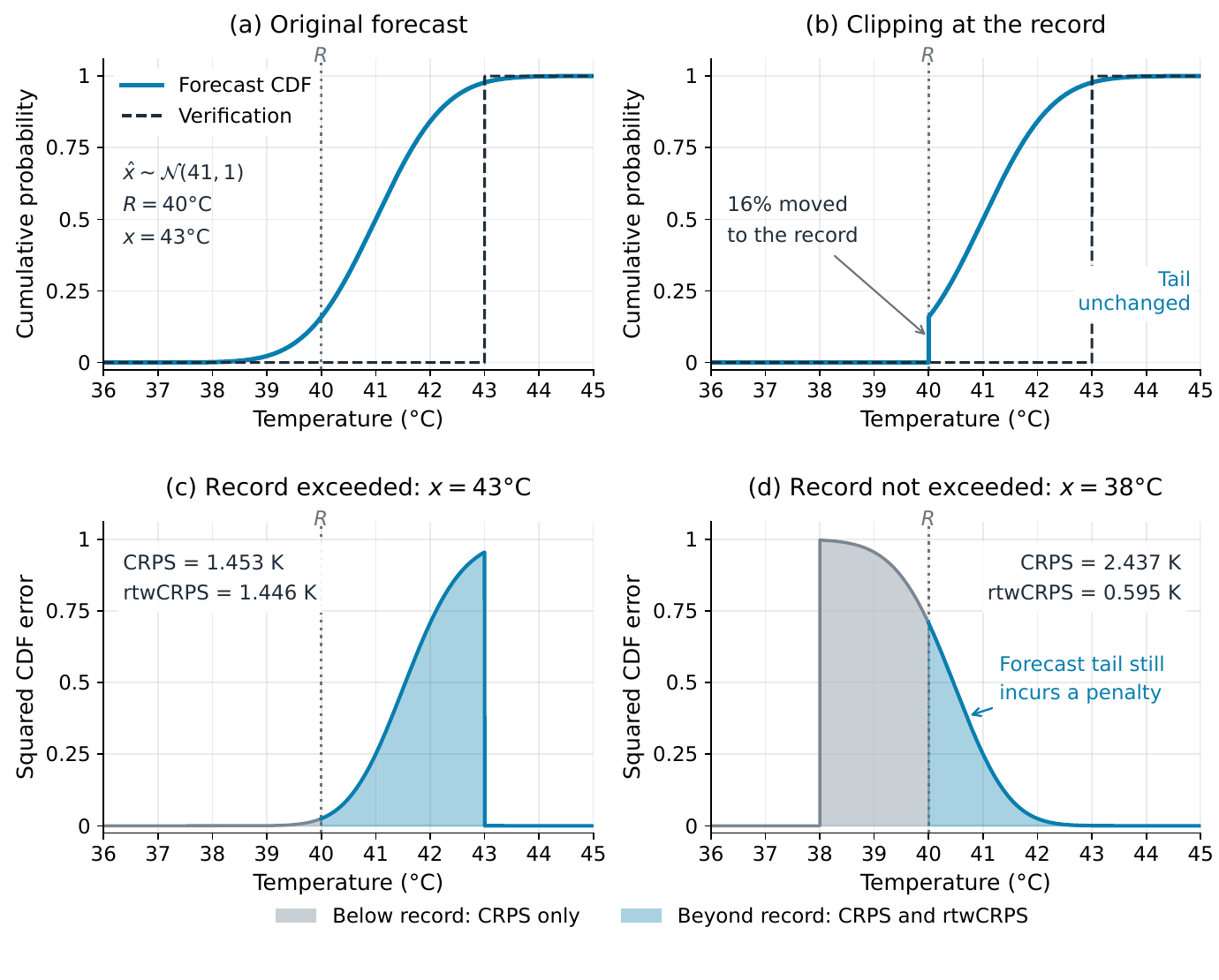}
    \caption{\textbf{Forecasting and evaluating a record-breaking
    extreme.}
    The forecast distribution is Gaussian with mean
    $41\,^{\circ}\mathrm{C}$ and standard deviation $1\,\mathrm{K}$;
    the historical record is $R=40\,^{\circ}\mathrm{C}$.
    (a) Forecast CDF and the step function corresponding to the
    verifying value $x=43\,^{\circ}\mathrm{C}$.
    (b) Clipping forecast and verifying values at $R$ collapses
    values below the record onto the boundary. Here, approximately
    16\% of forecast probability moves to $R$, while the upper
    tail remains unchanged.
    (c) Squared CDF error for the record exceedance.
    CRPS integrates the combined grey and blue areas; rtwCRPS
    retains only the blue area beyond $R$, equivalently computing
    CRPS after clipping.
    (d) The same forecast evaluated against
    $x=38\,^{\circ}\mathrm{C}$. Forecast probability beyond the
    record still incurs a penalty, retaining false alarms in
    the assessment. For this forecast, rtwCRPS is identical for
    all verifying values $x\leq R$.
    Raising the threshold reduces the integration
    region and therefore cannot increase rtwCRPS for a fixed
    forecast and verifying value.}
    \label{fig:crpsrtwcrpsschematic}
\end{figure}

	It measures the integrated squared difference between the forecast CDF and the step function associated with the observation (Figure~\ref{fig:crpsrtwcrpsschematic}a). Lower values indicate better forecasts. A proper score is minimized in expectation by reporting the true conditional distribution \citep{gneiting2007strictly}.
	
	We evaluate ordinary CRPS alongside two scores focused on the record tail. To account for differing ensemble sizes $M$, we use the fair CRPS estimator \citep{ferro2014fair}:
	
    \begin{equation}
		\fCRPS(\{\hat{x}_j\}_{j=1}^{M},x)
		=\frac{1}{M}\sum_{j=1}^{M}|\hat{x}_j-x|
		-\frac{1}{2M(M-1)}
		\sum_{j\ne l}|\hat{x}_j-\hat{x}_l|.
		\label{eq:fair}
	\end{equation}
	
    For upper extremes, record-threshold weighted CRPS restricts integration to values above $u_i(k)$:
    
	\begin{equation}
		\rtw_{u_i(k)}(F_i,x_i)
		=\int_{u_i(k)}^{\infty}
		[F_i(z)-\ind\{x_i\le z\}]^2\,\mathrm dz.
		\label{eq:rtw}
	\end{equation}
    
For lower extremes, the integral instead runs from $-\infty$ to $u_i(k)$. This applies the threshold-weighted CRPS of \citet{gneiting2011comparing} to local historical records and increasingly deep exceedances beyond them. Equivalently, it is CRPS after censoring both forecasts and observations at the threshold. Writing $u=u_i(k)$, the transformation is $v_u(x)=\max(x,u)$ for upper extremes and $v_u(x)=\min(x,u)$ for lower extremes. The fair ensemble estimate is

\begin{equation}
	\widehat{\rtw}_{u}(\{\hat{x}_j\}_{j=1}^{M},x)
	=\frac{1}{M}\sum_{j=1}^{M}|v_u(\hat{x}_j)-v_u(x)|
	-\frac{1}{2M(M-1)}
	\sum_{j\ne \ell}|v_u(\hat{x}_j)-v_u(\hat{x}_{\ell})|.
	\label{eq:frtw}
\end{equation}

The transformation moves all probability outside the selected tail to the threshold boundary, without renormalizing the tail (Figure~\ref{fig:crpsrtwcrpsschematic}b). The transformed distribution therefore holds information about both event probability and magnitude.
	
	Figure~\ref{fig:crpsrtwcrpsschematic}c,d illustrates why evaluation must keep both exceeding and non-exceeding cases. When a record occurs, insufficient forecast probability at the observed severity contributes to the loss. When no record occurs, forecast probability above the threshold still incurs loss, as shown by the blue area in panel~(d). The score thus accounts for missed records and false alarms, while focusing attention on the tail region of interest.
	
    For occurrence alone, we estimate the forecast event probability and define the observed event indicator as
    
    \begin{equation}
    	p_i(k)=\frac{1}{M}\sum_{j=1}^{M}
    	\ind\{\hat{x}_{ij}>u_i(k)\},
    	\qquad
    	o_i(k)=\ind\{x_i>u_i(k)\}
    \end{equation}
    
    for upper extremes. For lower extremes, the events are instead $\hat{x}_{ij}<u_i(k)$ and $x_i<u_i(k)$, respectively. The fair Brier score is then
    
    \begin{equation}
    	\widehat{\operatorname{BS}}_i(k)
    	=[p_i(k)-o_i(k)]^2
    	-\frac{p_i(k)[1-p_i(k)]}{M-1}.
    	\label{eq:brier}
    \end{equation}
	
	Both record scores are averaged over all eligible cases at every threshold. Threshold weighting preserves properness because the non-negative weight $\ind\{z\ge u_i(k)\}$  is specified from historical data without using the verifying outcome, in contrast to \eqref{eq:condcrps}. It focuses on the tail without rewarding systematic overstatement of extreme-event probabilities. However, rtwCRPS cannot distinguish distributions that differ only below the threshold, so ordinary CRPS remains necessary to assess the full distribution.
	
	Raising the lower integration boundary in equation~\eqref{eq:rtw} removes non-negative contributions from the score. In Figure~\ref{fig:crpsrtwcrpsschematic}c,d, this corresponds to retaining a smaller portion of the shaded area. A decrease in absolute rtwCRPS with depth therefore does not, by itself, indicate improved forecast performance. Instead, we assess comparative performance using IFS-ENS as the reference at each threshold by means of the skill score:
    
	\begin{equation}
		\operatorname{Skill}_S(k)
		=100\left[
		1-\frac{\overline S_{\rm AI}(k)}
		{\overline S_{\rm IFS}(k)}
		\right],
		\qquad
		\overline S(k)
		=\frac{\sum_i w_i S_i(k)}{\sum_i w_i},
		\label{eq:skill}
	\end{equation}

    with $S\in\{\widehat{\operatorname{CRPS}},\widehat{\operatorname{rtwCRPS}_{u}},\widehat{\operatorname{BS}}\}$ and  area weights $w_i=\cos(\phi_i)$, with $\phi_i$ the latitude of case $i$. Positive skill favours AI and negative skill favours IFS-ENS. An empirical climatology constructed from the same historical sample would by definition assign zero probability beyond its record and is unsuitable as a baseline. Consequently, increasing skill relative to IFS-ENS can reflect changes in either system's score.
	
	To identify \textit{where} score differences arise, we partition each average into contributions from observed exceedances and non-exceedances:
    
	\begin{equation}
		\overline S(k)
		=\frac{\sum_i w_i S_i(k)o_i(k)}{\sum_i w_i}
		+\frac{\sum_i w_i S_i(k)[1-o_i(k)]}{\sum_i w_i}.
		\label{eq:contributions}
	\end{equation}
    
	Both terms keep the all-case denominator and sum exactly to the primary score. For comparison, record-conditioned CRPS evaluates the ensemble distributions only where an exceedance occurs:
    
	\begin{equation}
		\overline{\CRPS}_{\rm cond}(k)
		=\frac{\sum_i w_i o_i(k)\,\fCRPS(\{\hat{x}_{ij}\}_{j=1}^{M},x_i)}
		{\sum_i w_i o_i(k)}.
		\label{eq:condcrps}
	\end{equation}
    
	Here the outcome-dependent weight lies outside the score. In Figure~\ref{fig:crpsrtwcrpsschematic}, this would retain the full grey and blue area for the record case in panel~(c), while excluding the non-record case in panel~(d). By contrast, rtwCRPS retains the blue area in both cases. Record conditioning therefore excludes false alarms on non-exceeding cases and can favour forecasts that assign excessive probability to extremes, an instance of the forecaster's dilemma \citep{lerch2017forecasters}. We use it as a descriptive diagnostic alongside the all-case proper scores.
	
	Finally, the occurrence ratio
    
	\begin{equation}
		Q(k)
		=\frac{\sum_i w_i p_i(k)}
		{\sum_i w_i o_i(k)}
		\label{eq:ratio}
	\end{equation}
    
	compares mean forecast exceedance probability with observed frequency. Values below one indicate an aggregate probability deficit and values above one a surplus. Agreement near one does not establish calibration, because opposing errors across cases can still cancel. We interpret $Q$ together with the Brier scores, record-conditioned CRPS and the exceedance and non-exceedance contributions to understand AI model improvements or deteriorations beyond the record.
	
	\section{Results} 
	\subsection{AI ensembles record skill}
	
\begin{figure}[htbp]
		\centering
		\includegraphics[width=\textwidth]{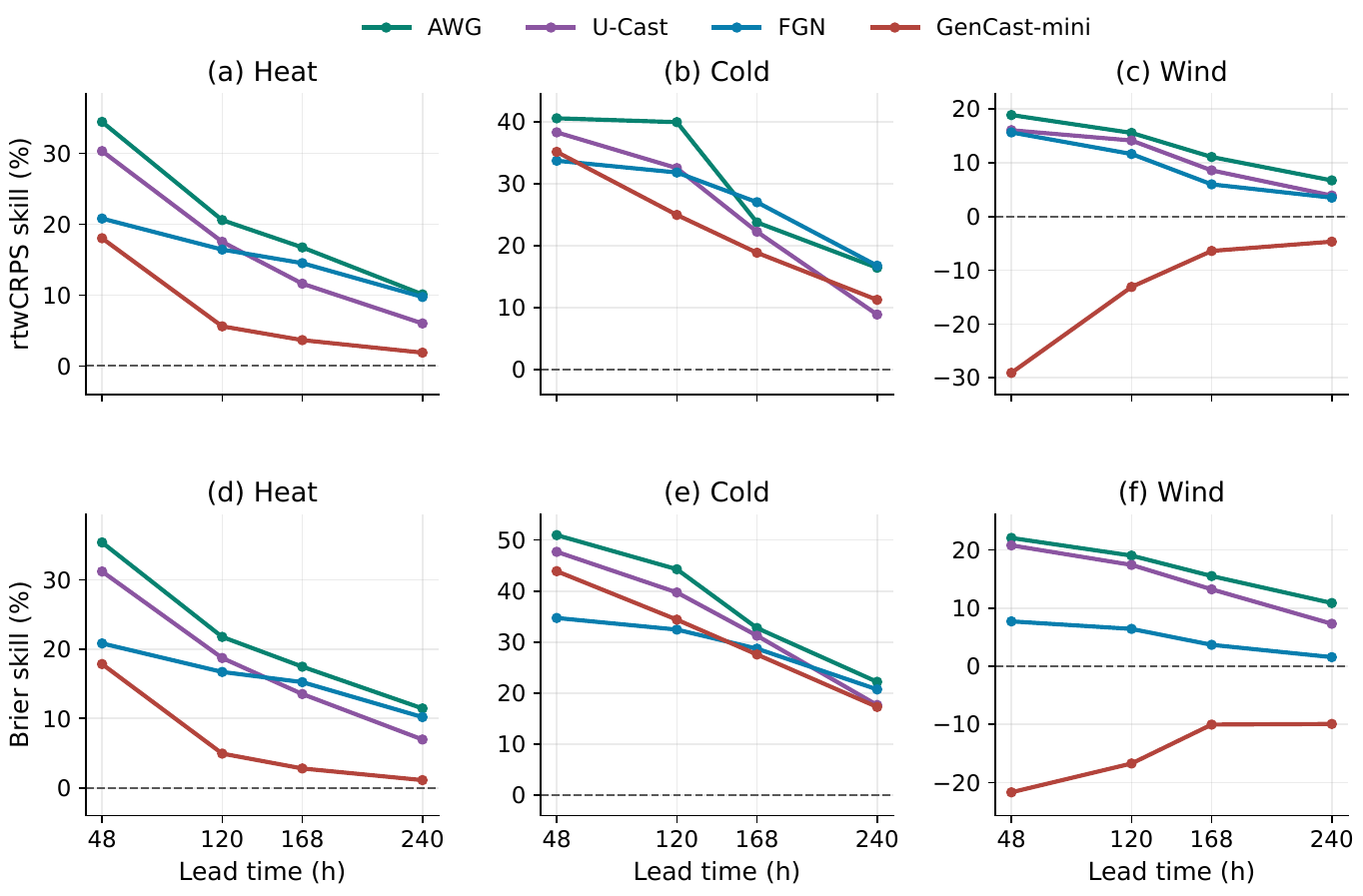}
		\caption{\textbf{rtwCRPS and Brier skill scores as a function of lead time.} Fair rtwCRPS skill (top) and Brier skill (bottom) relative to IFS-ENS at the 1979--2021 monthly record boundary. Positive values indicate better scores than IFS-ENS.}
		\label{fig:recordskill}
	\end{figure}

	At the historical record boundary ($k=0$), AWG, U-Cast and WN2 achieve lower rtwCRPS and Brier scores than IFS-ENS for heat, cold and wind at every evaluated lead from 48 to 240~h (Figure~\ref{fig:recordskill}). GenCast-mini also improves both temperature record scores at every lead, but performs worse for wind. Positive record skill is therefore shared across several native probabilistic AI systems, including the lightweight GenCast configuration, while the wind comparison distinguishes their capabilities. 
	
	At 48~h, rtwCRPS reductions across the four AI systems range from 18.0 to 34.4\% for heat and from 33.7 to 40.6\% for cold, with AWG providing the largest improvement in both cases. These advantages generally narrow with lead time: at 240~h, the corresponding ranges are 1.9--10.1\% and 8.9--16.8\%. For wind at 48~h, AWG, U-Cast and WN2 improve rtwCRPS by 18.9\%, 16.1\% and 15.7\%, respectively. GenCast-mini instead has 29.1\% higher rtwCRPS and 21.7\% higher Brier score than IFS-ENS, and remains worse in both wind scores at the longer leads.
	
	Performance at the record boundary does not simply follow ordinary forecast quality. All four AI systems improve all-case temperature CRPS, but AWG and U-Cast have respectively 19.5\% and 18.6\% higher all-case wind CRPS than IFS-ENS at 48~h, despite improving both record scores (Figure~\ref{fig:overall}). The full distribution and the record tail consequently give complementary assessments of the same forecasts.
	
	\subsection{Skill beyond the historical record}
	
\begin{figure}[htbp]
		\centering
		\includegraphics[width=\textwidth]{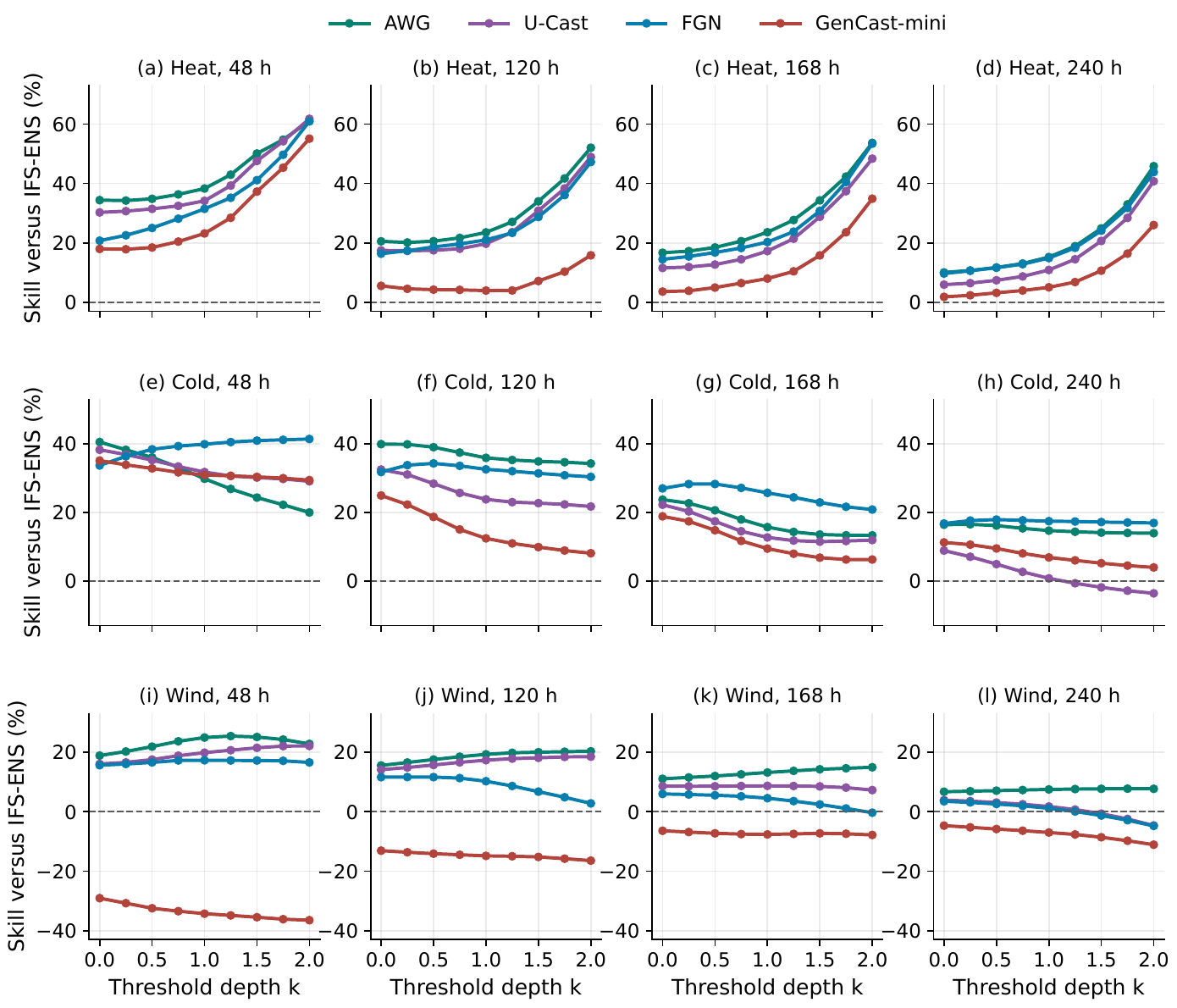}
		\caption{\textbf{Depth-based rtwCRPS skill across lead times.} Fair rtwCRPS skill relative to IFS-ENS, verified against ERA5 over land in 2022. Increasing $k$ moves the threshold beyond the historical record in units of local historical variability. All eligible cases enter the score at every threshold.}
		\label{fig:eradepth}
	\end{figure}

	The relative heat advantage generally strengthens as the evaluation threshold moves beyond the historical record (Figure~\ref{fig:eradepth}). At 48~h, rtwCRPS skill across the four AI systems increases from 18.0--34.4\% at $k=0$ to 55.1--61.7\% at $k=2$. At 240~h, the corresponding ranges are 1.9--10.1\% and 26.0--45.8\%. GenCast-mini retains positive rtwCRPS skill, although its gains are generally smaller than those of the other AI systems. Its Brier scores for $k \le 1.5$ become broadly comparable to IFS-ENS from 120~h onwards, while AWG, U-Cast and WN2 retain clearer advantages (Figure~\ref{fig:bsERA}).
	
	For cold, the AI advantage generally persists, but its dependence on depth varies between models. At 48~h, rtwCRPS skill decreases between $k=0$ and $k=2$ from 40.6\% to 20.0\% for AWG, from 38.3\% to 29.1\% for U-Cast, and from 35.2\% to 29.5\% for GenCast-mini. WN2 instead improves from 33.7\% to 41.5\%. AWG, WN2 and GenCast-mini retain positive cold rtwCRPS skill across all evaluated depths and leads. U-Cast falls below IFS-ENS only at the deepest 240-hour threshold ($-3.6\%$), while its Brier skill remains positive ($+5.7\%$).
	
	When it comes to wind, AWG improves both record scores throughout $0\le k\le2$ at all four leads. U-Cast and WN2 retain positive rtwCRPS skill throughout this range at 48~h, but lose their advantage at deeper thresholds for longer leads. At 240~h and $k=2$, their rtwCRPS skills are $-4.6\%$ and $-4.8\%$, respectively. WN2 also has negative Brier skill ($-20.8\%$), whereas U-Cast retains a small improvement ($+2.6\%$). GenCast-mini performs worse than IFS-ENS in both wind scores throughout the evaluated range, with its 48-hour rtwCRPS skill declining from $-29.1\%$ at the historical boundary to $-36.5\%$ at $k=2$.
	
	As expected, the number of observed exceedances falls sharply with depth: at $k=2$, only 28 heat, 203 cold and 200 wind grid-cell--time exceedances remain (Table~\ref{tab:counts}), and neighbouring cases may belong to the same meteorological events (Figure~\ref{fig:supp_r_depth}). Although non-exceedances contribute to the proper scores, the deepest thresholds provide limited direct evidence about performance on realized record-shattering events; comparisons at the deepest thresholds therefore warrant caution.
	
	\subsection{Where does skill come from?}
	
\begin{figure}[htbp]
		\centering
		\includegraphics[width=\textwidth]{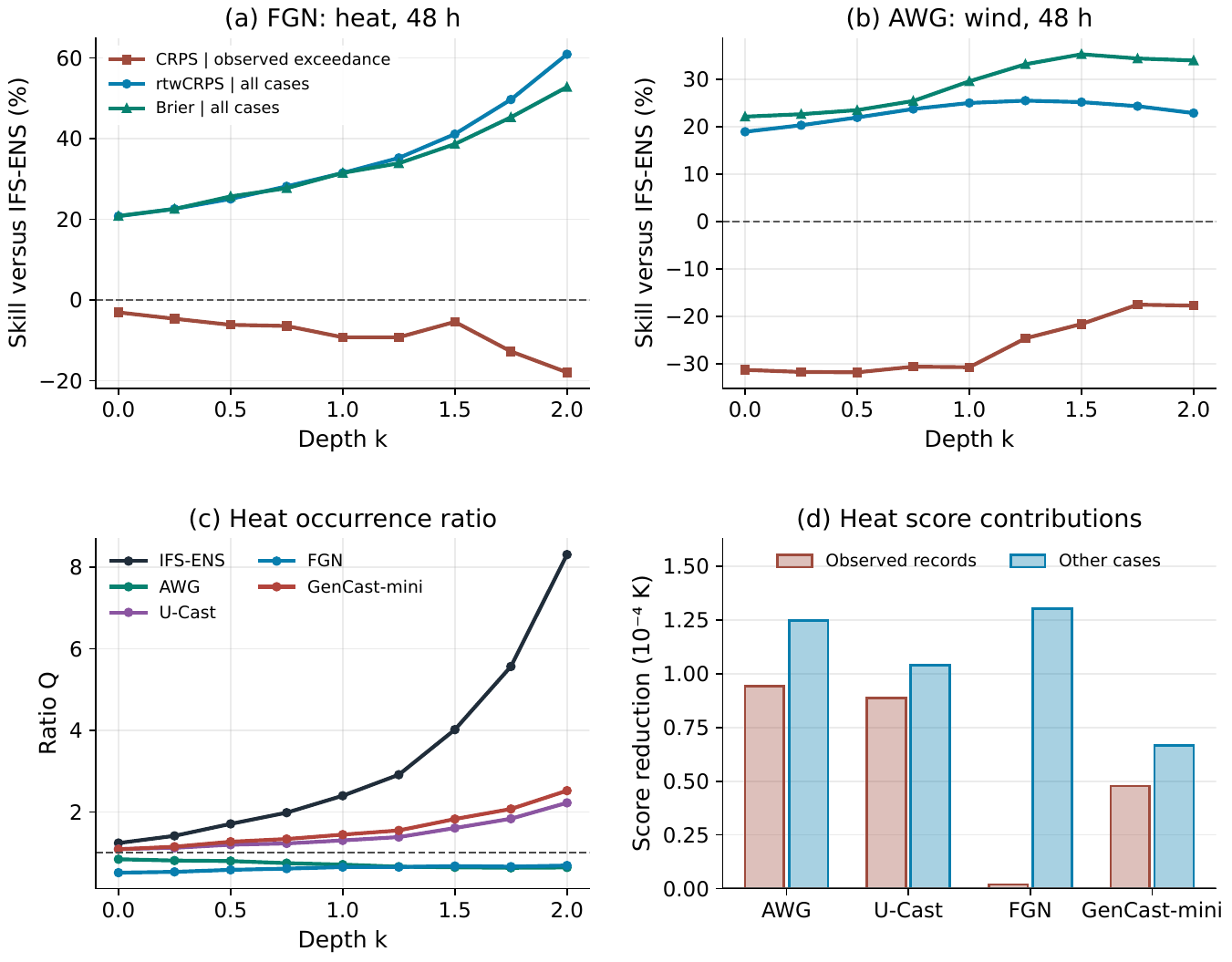}
		\caption{\textbf{Proper versus record-conditioned score rankings and score contributions.} 48-hour forecasts verified against ERA5. (a,b) WN2 heat and AWG wind illustrate different rankings from record-conditioned CRPS and all-case record scores. (c) Mean forecast heat-exceedance probability divided by observed frequency for all five systems; unity indicates aggregate frequency agreement. (d) Observed-exceedance and non-exceedance contributions to the fair heat rtwCRPS reduction relative to IFS-ENS at $k=0$ as in \eqref{eq:contributions}. Positive contributions favour AI. WN2's improvement arises almost entirely on non-exceedances, whereas AWG, U-Cast and GenCast-mini reduce both contributions.}
		\label{fig:selection}
	\end{figure}

	All-case record scores and record-conditioned CRPS can rank the same forecasts differently (Figure~\ref{fig:selection}a,b). For heat at 48~h and $k=0$, WN2 improves rtwCRPS by 20.8\% relative to IFS-ENS, yet its conditional CRPS is 3.1\% worse (1.345 vs 1.305~K). The contrast grows for deeper events: at $k=2$ (28 land events), WN2 improves rtwCRPS and the Brier score by 60.9\% and 52.8\%, while its conditional CRPS is 17.9\% worse. GenCast-mini shows the same pattern, with 55.1\% better rtwCRPS and 20.3\% worse conditional CRPS. Wind behaves similarly: at 48~h and $k=0$, AWG and U-Cast improve both all-case scores, but their conditional CRPS is 31.3\% and 26.7\% higher.
	
    The two scores differ in which cases they evaluate and which part of the distribution they penalise. Conditional CRPS scores the full distribution, but only on observed exceedances; rtwCRPS scores only the tail, but on all cases. The contribution analysis in Equation~\eqref{eq:contributions} shows where rtwCRPS gains arise. For heat at 48~h and $k=0$, about 99\% of WN2's improvement comes from non-exceedances, while its loss on observed exceedances barely changes. AWG, U-Cast and GenCast-mini reduce both contributions, with non-exceedances accounting for 54\% to 58\% of the gain (Figure~\ref{fig:selection}d).
	
	At $k=2$, the heat improvements rely even more on non-exceedances. For AWG, WN2 and GenCast-mini, reduced non-exceedance loss outweighs higher loss on the observed exceedances; for U-Cast, it accounts for approximately 98\% of the improvement. Cold provides a contrasting result. At 48~h and $k=2$, all four AI systems reduce the observed-exceedance contribution. WN2's mean rtwCRPS on those cases is 0.379~K, compared with 0.737~K for IFS-ENS. This reduction outweighs slightly greater non-exceedance loss and produces a 41.5\% overall improvement. For GenCast-mini, observed exceedances supply about 86\% of cold improvement. Favourable tail scores therefore do not reflect a single mechanism.
	
	Occurrence ratios separate comparative skill from probability agreement. For heat at 48~h and $k=2$, IFS-ENS forecasts records far too often ($Q = 8.31$). U-Cast and GenCast-mini forecast it too often ($Q = 2.22$ and $2.52$), and AWG and WN2 too rarely ($Q = 0.64$ and $0.68$; Figure~\ref{fig:selection}c), yet all four AI systems improve both heat scores. Wind shows similar probability excesses: at 240~h and $k=2$, $Q$ is 5.32 for WN2 and 6.85 for GenCast-mini (Figure~\ref{fig:qERA}). Relative skill, accurate record frequencies and performance on realized exceedances are thus distinct properties, and neither positive skill nor $Q$ close to one establishes conditional tail calibration.
	
	\subsection{Sensitivity to the verification reference}
	
	Changing the common verification reference alters score rankings substantially (Figures~\ref{fig:bsHRES},~\ref{fig:truthswitch}). For temperature, against HRES-fc0 and compared to IFS-ENS, at 48~h, all-case CRPS is 58.4\% higher for AWG, 43.2\% higher for U-Cast and 60.7\% higher for GenCast-mini. WN2 instead achieves a 30.6\% reduction. This contrast is consistent with a role for training and initialization references: WN2 is fine-tuned and initialized using HRES-fc0, whereas the other AI systems use ERA5. The benchmark does not isolate this effect from other differences between the models.

    For heat records, AWG, U-Cast, and GenCast-mini rtwCRPS skills are generally negative at $k = 0$ and improve monotonically to positive values with increasing depth, where the increase is largest for the shortest lead time. One exception is GenCast-mini at 120~h lead, which does not improve rtwCRPS over IFS-ENS regardless of threshold depth. Contrary to all-case CRPS, the trend in deeper heat advantage thus seems to hold under most conditions when swapping the verification data.
	
\begin{figure}[htbp]
		\centering
		\includegraphics[width=\textwidth]{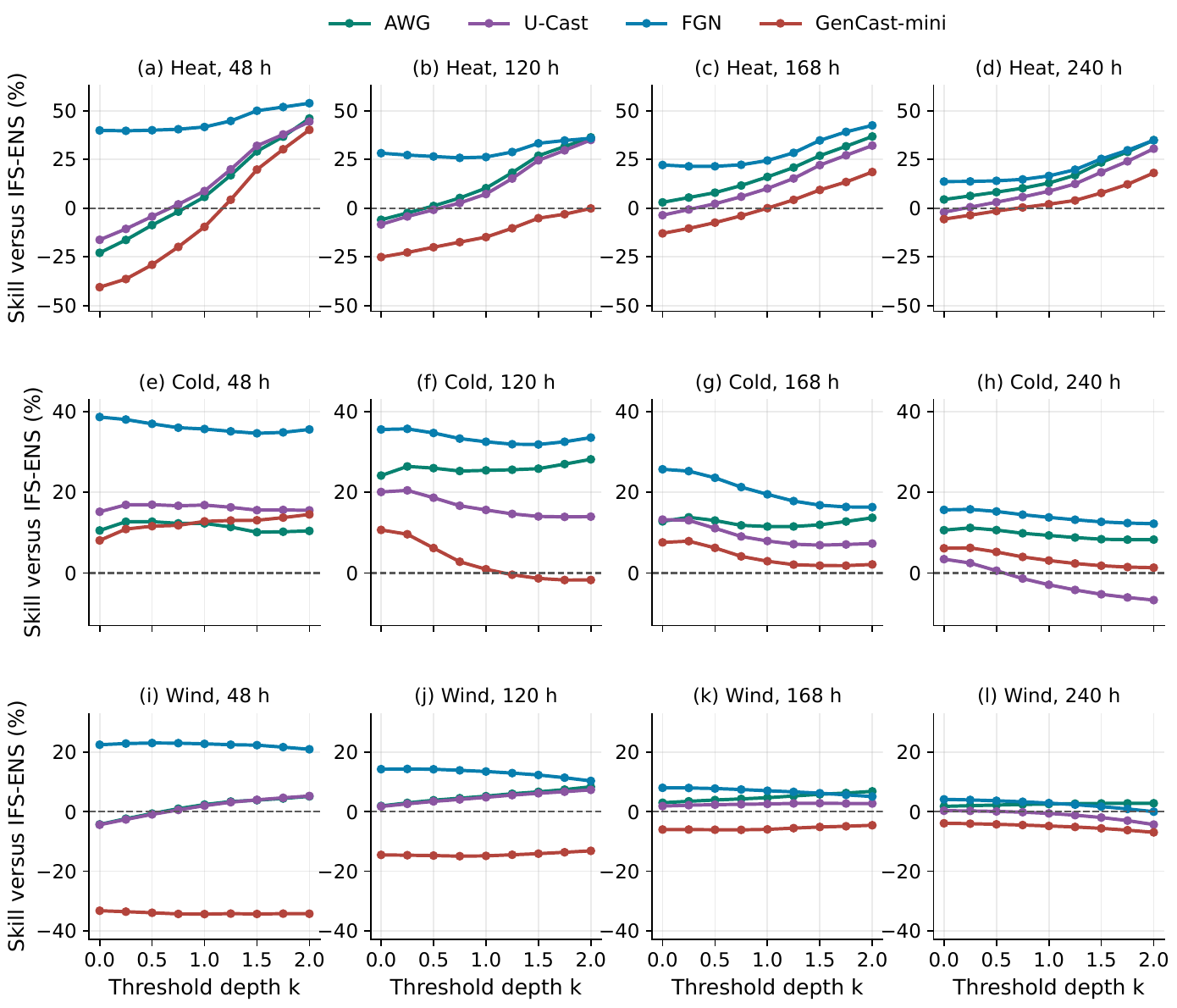}
		\caption{\textbf{Depth-based skill when verifying against HRES-fc0.} Fair rtwCRPS skill against HRES-fc0, using the same 2022 land sample and ERA5 historical records and scales for heat (top), cold (middle) and wind (bottom).}
		\label{fig:hresdepth}
	\end{figure}

	For cold records, AWG and WN2 improve both scores throughout the evaluated range, whereas U-Cast and GenCast-mini show near-zero and even negative rtwCRPS skill at certain lead time--threshold combinations. For wind, there is a pattern of positive skill for WN2, neutral results for U-Cast and AWG, and negative skill for GenCast-mini---a spread which diminishes with increasing lead time (Figure~\ref{fig:hresdepth}).

	The reference switch changes which cases exceed the historical thresholds, but not the confounding ERA5 records and scales. HRES-fc0 identifies 3,514 heat crossings compared with 5,129 in ERA5, but more cold crossings (2,275 versus 1,665) and wind crossings (5,838 versus 2,971; Table~\ref{tab:counts}). This shows that reference choice affects the observed event population as well as forecast errors.
	
	\subsection{A revised record benchmark of deterministic forecasts}
	
	The sensitivity of the results above to the choice of verification dataset motivates revisiting the deterministic benchmark of \citet{zhang2026recordextremes}. We compare HRES, GraphCast, Pangu-Weather and FuXi forecasts for 2020, using ERA5 historical records from 1979--2019. We vary the references used to identify record-breaking events and calculate forecast errors independently between ERA5 and HRES-fc0. The comparison below focuses on 48~h; the full design, figures, tables and additional leads are presented in Supplement~\ref{sec:detmethods}.
	
	We find that here, too, temperature rankings depend strongly on the error reference. On ERA5-selected heat and cold records, all three AI models have lower RMSE than HRES against ERA5, whereas HRES performs best against HRES-fc0 (Supplementary Figure~\ref{fig:reference2020}; Table~\ref{tab:detRMSE}). This sensitivity is already present in all-case errors, and temperature F1 rankings also change with the event reference (Table~\ref{tab:detOccurrence48}). Even when narrowing down to the subset of cases where both an AI model and HRES correctly forecast a record-breaking heat event, the RMSE reversal persists (Table~\ref{tab:detJoint48}). Wind shows a more persistent intensity deficit on observed records: conditioning on record exceedances, HRES has lower conditional RMSE under every reference combination and evaluated lead, despite competitive all-case AI errors. Occurrence comparisons and the jointly forecast record control yield less consistent rankings (Tables~\ref{tab:detOccurrence48} and~\ref{tab:detJoint48}), so HRES's advantage is specific to the intensity assessment on observed exceedances.
	
	A further note of caution concerns the increasing underprediction seen at greater observed depths. Selecting progressively more extreme outcomes can produce increasingly negative bias even for an ideal conditional-mean forecast \citep{lerch2017forecasters}; the derivation and its qualifications are given in Supplement~\ref{sec:detconditional}. The benchmark therefore identifies reference-sensitive temperature comparisons and more persistent wind errors on observed records, without establishing any general structural inability to extrapolate.
	
	\FloatBarrier
	
	\section{Discussion} 
	
	Our results show that several native probabilistic AI systems retain comparative skill beyond local historical records. The evidence is clearest for temperature: gains are shared across models at the record boundary, and the relative heat advantage often strengthens at greater depth and persists under common HRES-fc0 verification (Figures~\ref{fig:recordskill}, \ref{fig:eradepth} and~\ref{fig:hresdepth}). For cold, the AI advantage generally holds, with some exceptions at depth and longer leads. Wind shows less consistent gains and clearer differences between systems, including the persistent deficit of GenCast-mini over land. WN2 over-forecasts land wind records, its Brier skill score turns negative at depth and long leads, and AWG is the only model that stays positive at 240~h. These patterns support a differentiated assessment of probabilistic AI forecasts across variables and models. They also show that promising record scores can be achieved by the native forecast distributions, without additional statistical recalibration.
	
	Which score is used changes what the comparison shows. Record-conditioned CRPS describes the full predictive distribution on cases where a record occurred, whereas rtwCRPS evaluates the retained tail across all eligible cases, including those without an exceedance. The two scores rank models differently because they score different cases and different parts of the distribution. Outcome selection also affects deterministic intensity diagnostics. Even an ideal conditional-mean forecast underpredicts, on average, outcomes selected from its predictive upper tail; stronger underprediction can arise simply by raising the selection threshold \citep{lerch2017forecasters}. Growing underprediction with depth is therefore not, by itself, evidence of a structural forecast ceiling. The derivation is given in Supplement~\ref{sec:detconditional}, together with the qualifications needed when depth curves pool different forecast situations.
	
	Favourable all-case scores, in turn, do not establish that the most extreme outcomes are well predicted, or that tail probabilities are calibrated. Much of the heat improvement arises on non-exceedances (Figure~\ref{fig:selection}), while the cold results also show gains on realized extremes. Assigning little probability to events that do not occur is a real part of rare-event forecasting, but it provides different evidence from improving predictions when records are broken. The occurrence ratios further show that better relative scores can coexist with aggregate probability deficits or surpluses. A growing advantage with depth also depends on the behaviour of IFS-ENS and does not imply that deeper extremes become easier to forecast. Score contributions, occurrence diagnostics and conditional errors therefore help explain the gains without substituting for an assessment of conditional calibration.
	
	The deterministic revisit sharpens the distinction between temperature and wind. Temperature rankings can reverse when only the error reference changes, with forecasts and selected events held fixed (Figure~\ref{fig:reference2020}; Table~\ref{tab:detRMSE}). This sensitivity is already present in all-case errors, and the heat reversal persists when both models forecast a record (Table~\ref{tab:detJoint48}). It therefore extends beyond selection on observed extremes. For wind, HRES retains smaller intensity errors on observed-record subsets at the historical boundary across the reference combinations considered. Yet this advantage does not extend to every selection rule: pairwise comparisons on jointly forecast wind records favour the AI models against ERA5, and occurrence comparisons involve different precision--recall trade-offs. Our coarser-grid revisit of \citet{zhang2026recordextremes} thus identifies limitations whose interpretation depends on the variable, reference and selection rule; it does not isolate their structural cause.
	
	Verifying all systems against the same reference makes them comparable, while uncertainty in those references remains. Analysis and observation errors, together with dependence between forecasts and their verifying analyses, can affect scores and rankings \citep{ferro2017measuring,bessac2021forecast,ramon2024uncertainties,caron2025reliability}. Shared errors may also survive a reference switch, as illustrated by the surface-temperature artefacts reported by \citet{christensen2026error}. Agreement across ERA5 and HRES-fc0 therefore supports robustness to these reference choices, while disagreement does not establish which is closer to the physical atmosphere. Moreover, HRES-fc0 exceedances of an ERA5 historical threshold are crossings between products, rather than records within a homogeneous HRES-fc0 history. Because the records and scales come from ERA5, models trained on ERA5 share the climate that defines the boundary, whereas IFS-ENS is possibly penalised for its different model climate. Part of the relative advantage, especially its growth with depth for heat, might reflect this mismatch rather than better forecasts of the extremes themselves. Instead of either analysis, station measurements can serve as independent verification (e.g. \citealt{bremnes2024evaluation}; \citealt{gupta2026mausam}). Record benchmarks should explicitly distinguish the historical dataset defining the boundary, the reference selecting events and the reference measuring error. Ordinary scores alongside tail scores then help distinguish general reference sensitivity from deficiencies concentrated on extremes.
	
	The historical boundary also defines the scope of the extrapolation claim. A new local record may resemble conditions a globally trained model has seen elsewhere, and an unfamiliar atmospheric state need not break a record. A common historical threshold can also occupy a different position relative to the training extremes of models fitted on different periods. Differences in architecture, initialization and sampling further complicate attribution; WN2's more recent training data and HRES-fc0 fine-tuning are relevant examples. The present results compare available forecasting systems beyond a reproducible local boundary, without isolating a causal benefit of probabilistic training.

    Complementary evidence comes from the potential-score approach of \citet{biegert2026weighted}, building on \citet{gneiting2026potential}. Their threshold-weighted potential CRPS assesses distributions fitted to deterministic predictors through IDR/EasyUQ \citep{henzi2021idr}. This asks how much useful information about extremes a predictor contains after a common form of recalibration, which can differ from the quality of its raw amplitudes. Our native-ensemble comparison instead evaluates the distributions supplied by the forecasting systems themselves. Because potential scores use an optimal fit on the evaluation sample, the resulting gains require separate training and verification before they can be interpreted as achievable postprocessing performance on future records.
	
	The native-ensemble evidence covers one year on a 1.5$^\circ$ land grid using twice-daily instantaneous values. It does not resolve local peaks, wind gusts, event persistence or spatial coherence. Few exceedances remain at the deepest thresholds, and spatial and temporal dependence further reduce the effective sampling size (Table~\ref{tab:counts}). The comparisons are descriptive for this sample; small score differences warrant caution. Record depth is also a standardized historical distance that includes climate drift. Positive skill here denotes improvement over IFS-ENS and does not establish value relative to climatology or for a particular decision. Longer evaluations across years, regions and event definitions are needed to assess how consistently the observed advantages translate into useful forecasts of consequential extremes.
	
	Progress also depends on access to forecasts that permit these comparisons. Alongside current verification initiatives \citep{brocker2026verification,mctaggartcowan2026wpmip}, we encourage open releases of member-level probabilistic forecast archives with compatible variables, dates and leads, supported by clear documentation of training periods, initialization and ensemble generation. Open model weights and inference code are valuable, but accessible forecast archives are what make harmonized evaluation practical and reproducible. Comparing fixed forecast samples against multiple analyses and independent observations would help assess reference dependence and distinguish uncertainty in the historical boundary from uncertainty in subsequent exceedances. These resources would support more reliable conclusions about where probabilistic AI forecasts add value and where their extreme tails still need improvement.
	
	\section{Conclusion} 
	
	Native probabilistic AI ensembles can outperform IFS-ENS beyond local historical records. In our 2022 land benchmark, all four systems achieve lower temperature record scores against ERA5 at the historical boundary. Their relative heat advantage generally grows at deeper thresholds and largely persists under HRES-fc0 verification, while wind results are more mixed. These findings challenge a general claim that AI models cannot forecast beyond local historical records.
	
	Some improvements come mainly from lower false-alarm penalties, and forecast probabilities can still overstate or understate record frequency. Our deterministic comparison also shows that temperature rankings can reverse when the verification reference changes. Models should therefore be compared against the same reference, using scores that account for both missed records and false alarms. Evaluations should also examine predictions on the records that occur and whether events assigned a given probability occur at that frequency. Openly sharing ensemble forecasts would help others repeat and extend these comparisons.

    \section*{Acknowledgments}

    B.V.V., C.D., D.G.M., S.L., and T.M. conceived and designed the study. B.V.V., C.D. and T.M. conducted the experiments, analyzed the results and wrote the manuscript. All authors revised the manuscript.
    
    The authors acknowledge the support of the European Research Council (ERC) via the HEAT Consolidator grant (101088405). The computational resources and services used in this work were provided by the VSC (Flemish Supercomputer Center), funded by the Research Foundation, Flanders (FWO), and the Flemish Government. We thank Nusret Ipek for letting us use the BIOBOT machinery.
	
    \section*{Data and code availability}
	All code, results and manuscript materials are available in the ProbEx repository at \url{https://github.com/Cas-Dec/probex}. Data sources and access are described in Sections~\ref{sec:groundtruth} and~\ref{sec:probforecasts}.

	\clearpage
	\begingroup
	\small
	\setlength{\bibsep}{3pt}
	\bibliographystyle{plainnat}
	\bibliography{references}
	\endgroup
	\clearpage
	\appendix
	\renewcommand{\thesection}{S\arabic{section}}
	\setcounter{section}{0}
	\renewcommand{\thefigure}{S\arabic{figure}}
	\setcounter{figure}{0}
	\renewcommand{\thetable}{S\arabic{table}}
	\setcounter{table}{0}
    
	\section*{Supplementary material}

    \section{New records set in 2022}
    \label{sec:supp_records}

\begin{figure}[htbp]
		\centering
		\includegraphics[width=\textwidth]{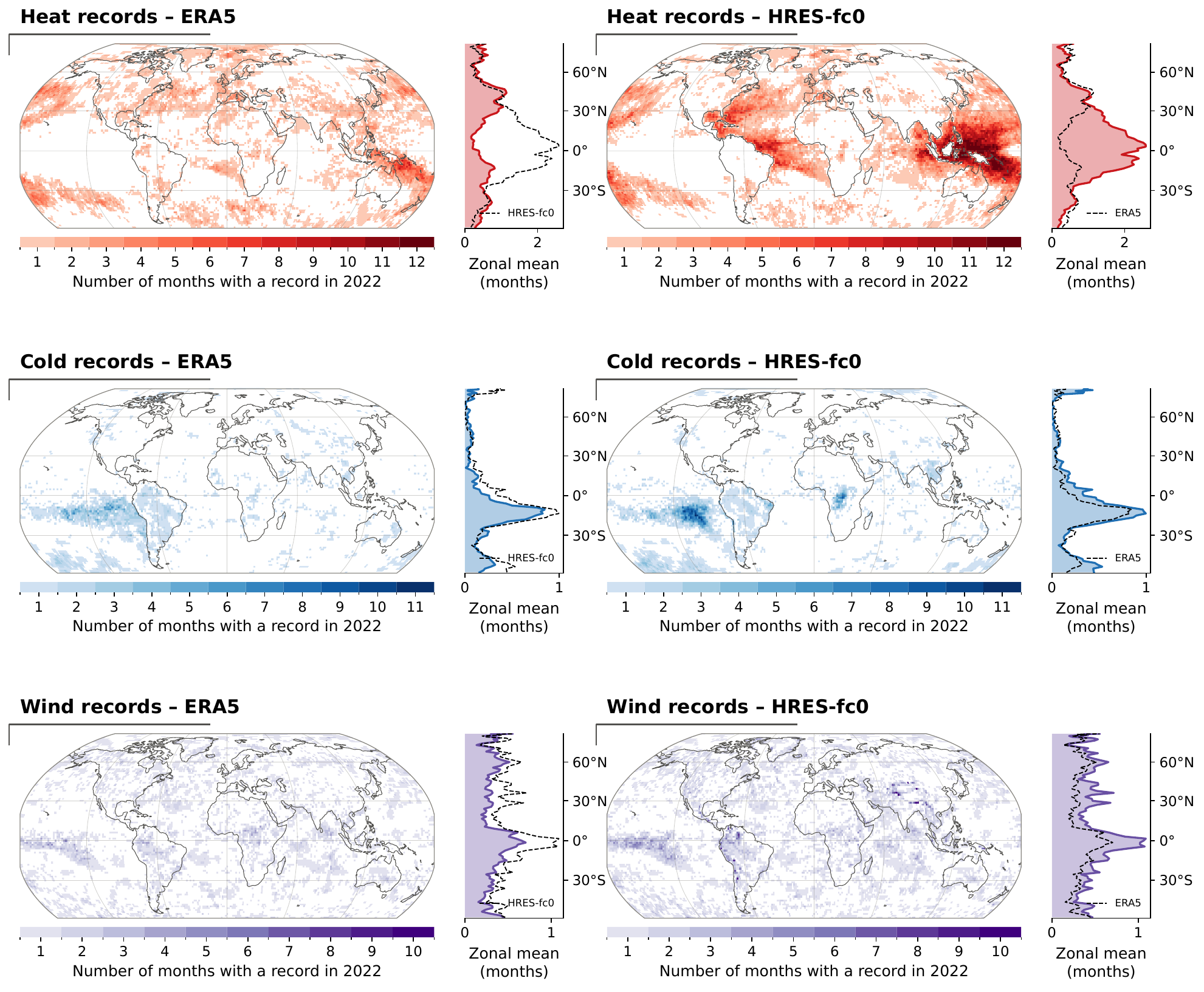}
		\caption{\textbf{ERA5 versus HRES-fc0 monthly records in 2022.} For each grid cell, the figure shows the number of months in the year 2022 that experienced record-breaking heat (top), cold (middle) and wind (bottom) for both ERA5 (left) and HRES-fc0 (right) against 1979--2021 historical ERA5 conditions.}
		\label{fig:supp_records}
	\end{figure}

\begin{figure}[htbp]
		\centering
		\includegraphics[width=\textwidth]{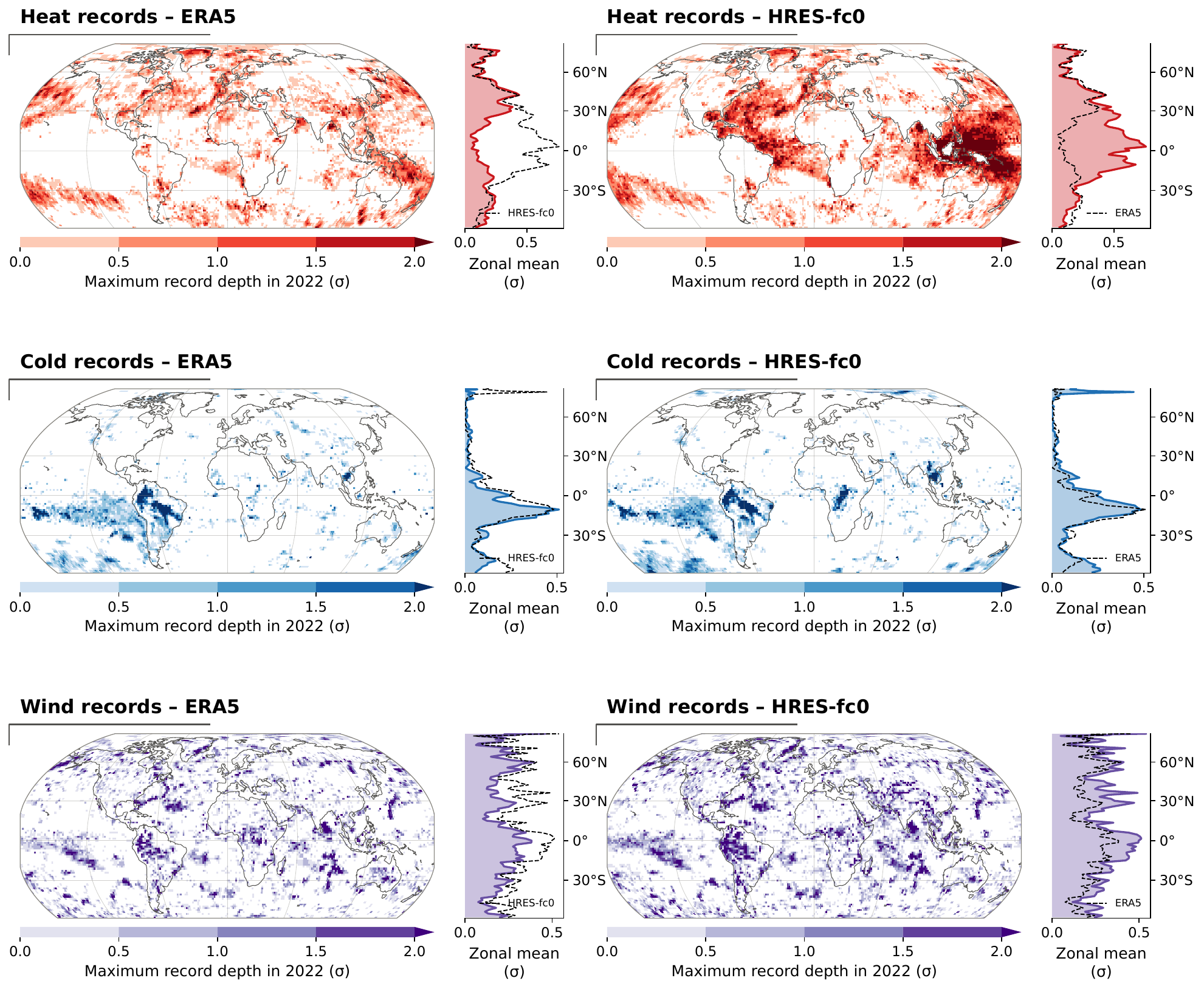}
		\caption{\textbf{ERA5 versus HRES-fc0 max record depths in 2022.} For each grid cell, the figure shows the maximum record depth in the year 2022 for heat (top), cold (middle) and wind (bottom) for both ERA5 (left) and HRES-fc0 (right) against 1979--2021 historical ERA5 conditions.}
		\label{fig:supp_r_depth}
	\end{figure}

    Figures~\ref{fig:supp_records} and~\ref{fig:supp_r_depth} show, for every grid cell, the number of months in 2022 that broke the 1979--2021 ERA5 monthly record and the deepest record of the year, using ERA5 or HRES-fc0 as verification. Both datasets agree on the large-scale picture and on most individual record events. Of the ERA5 record months, 73\% (heat), 56\% (cold), and 79\% (wind) are also flagged by HRES-fc0. Where both flag a record, the record depths correlate well for cold ($r=0.79$) and wind ($r=0.72$) extremes and moderately for heat extremes ($r=0.40$), with median depth differences below $0.2\sigma$. HRES-fc0, however, flags considerably more records: $1.9\times$ as many heat record months, $1.4\times$ as many cold and wind record months, and $7\times$ as many grid cells with a heat record deeper than $2\sigma$.

    These differences mostly come from the model climatologies, not from the 2022 weather itself. ERA5 is produced with a fixed IFS version (CY41r2, $\approx$31\,km) for its whole period \citep{hersbach2020era5}. HRES-fc0 was produced throughout 2022 with the much newer cycle CY47r3 at $\approx$9\,km resolution \citep{ecmwf2021cy47r3}. Because records and scales are defined from ERA5, any systematic offset between the two analyses acts as a shift of the record threshold. The effect is largest where the interannual variability of monthly extremes is small. Over tropical oceans ($10^\circ$S--$10^\circ$N), HRES-fc0 monthly maxima are on average about $1\sigma$ warmer than ERA5's. As a result, HRES-fc0 sets heat records almost every month over the tropical Atlantic, where ERA5 sets few. Over tropical land, HRES-fc0 maxima are about $1\sigma$ cooler, so some continental heatwaves that are clear in ERA5 (India--Pakistan in March, the Yangtze basin in July--August, the western United States in September) are weak in HRES-fc0.

    The ERA5 records themselves follow known features of 2022. Heat records are spread over land and ocean roughly in proportion to their area (21\% of the summer record depth falls over land, which covers 28\% of the domain). They are concentrated in the subtropics and lower mid-latitudes. The scarcity of equatorial records is consistent with 2022 being the third consecutive La Ni\~na year \citep{wmo2023state}. La Ni\~na cools the central and eastern equatorial Pacific and warms the western Pacific warm pool \citep{mcphaden2006enso}, as seen in the persistent records in the Coral Sea. In addition, the 1979--2021 baseline contains the very high tropical maxima of the 1997/98 and 2015/16 El Ni\~no events. Over the mid-latitude oceans, many records lie among the subtropical western boundary currents, which have warmed two to three times faster than the global-mean ocean surface \citep{wu2012enhanced}. In 2022, 58\% of the ocean surface experienced at least one marine heatwave \citep{wmo2023state}. The main land records match the major heatwaves of the year, such as India and Pakistan in March \citep{zachariah2023attribution}, western Europe including the first 40$^\circ$C in the United Kingdom in July \citep{wmo2023state}. Cold records are much rarer and are concentrated in South America and the adjacent south-eastern Pacific. Over the ocean, they are consistent with La Ni\~na and with the multidecadal cooling trend of the south-eastern Pacific \citep{seager2019strengthening}. Wind records show no comparably clear spatial structure; December contributes with the largest share of record area in both datasets.

    \newpage
	\section{Deterministic record benchmark}
	\label{sec:detmethods}\label{sec:detresults}
	
	\subsection{Data and verification design}
	The deterministic benchmark uses raw WeatherBench~2 HRES, GraphCast, Pangu-Weather and FuXi forecasts at 1.5$^\circ$ for 2020. GraphCast and Pangu are the ERA5-initialized products. The land mask, latitude restriction and 00/12~UTC verification hours follow the native-ensemble benchmark. Forecasts at 48, 120 and 240~h are matched to the same 686 valid times, from 11 January at 00~UTC to 18 December at 12~UTC, giving 4,649,708 land grid-cell--time pairs per model and lead.
	
	Historical records follow equation~\eqref{eq:records}, using ERA5 during 1979--2019 rather than 1979--2021. Local scales are the sample standard deviations of the corresponding 41 annual monthly extrema. The experiment revisits the question posed by \citet{zhang2026recordextremes} on the coarser grid used by \citet{biegert2026weighted}. Differences in resolution, historical baseline, sampling hours, model variants and matched dates preclude an exact replication claim.
	
	We keep the ERA5 historical boundary fixed and vary the event selector $E$ and error reference $T$ independently:
	\begin{equation}
		(E,T)\in\{\mathrm{ERA5},\mathrm{HRES\text{-}fc0}\}
		\times\{\mathrm{ERA5},\mathrm{HRES\text{-}fc0}\}.
		\label{eq:permutations}
	\end{equation}
	Changing $T$ at fixed $E$ isolates error-reference sensitivity on unchanged selected cases. Changing $E$ at fixed $T$ isolates event-selection sensitivity. All four combinations use the same eligible sample, land mask, weights, historical thresholds and local scales.
	
	\subsection{Intensity and occurrence diagnostics}
	Conditional bias and RMSE use cumulative subsets defined by the $E$-based physical record margin $D_i>v$ or standardized depth $d_i>k$, with forecast errors calculated against $T$. All-case RMSE is calculated before event selection and therefore depends on $T$ only. The conditional subsets shrink as the cutoff increases. Cold errors are reflected so that negative severity bias denotes insufficient extremeness for heat, cold and wind alike.
	
	Occurrence diagnostics use area-weighted precision, recall and F1 at the issued, unshifted record threshold. Forecast crossing counts are reported separately as unweighted cell--time counts. Precision--recall sweeps shift each forecast towards greater severity by $g$ times its local calendar-month forecast standard deviation, with $g=-1.5,-1.45,\ldots,1.5$; the shift is subtracted for cold. This is the same framework for the decision boundary as in \cite{zhang2026recordextremes}. Importantly, this standard deviation is estimated from the 2020 forecast sample and differs from the previously issued historical scaled $s_i$. The reference events remain fixed. 
	
	The joint-forecast control selects cases where a particular AI model and HRES both predict a record, independently of the verifying outcome. Each pair is then evaluated against both references on that same subset. Subsets differ between pairs, so comparisons are within each pair. 
	
	Finally, with selected cases and weights fixed, changing the error reference adds the same mean reference offset to every model's signed bias. Between-model signed-bias differences are therefore unchanged, even when absolute-bias or RMSE rankings reverse. 
	
	\subsection{Reference agreement and event support}
	The references identify substantially different sets of record exceedances (Table~\ref{tab:overlap2020}). ERA5 identifies more heat crossings, whereas HRES-fc0 identifies more wind crossings. These differences concern exceedances of the same ERA5 historical boundary and establish disagreement about event occurrence without identifying which reference is physically more accurate.
	
	\begin{table}[htbp]
		\centering\small
		\caption{Agreement between the two 2020 references at the ERA5-defined record boundary. F1 is the binary-event overlap, $2N_{\rm both}/(N_{\rm ERA5}+N_{\rm HRES-fc0})$.}
		\label{tab:overlap2020}
		\begin{tabular}{lrrrrrr}
			\toprule
			Extreme & ERA5 & HRES-fc0 & Both & ERA5 only & HRES-fc0 only & F1\\
			\midrule
			Heat & 7,754 & 4,102 & 2,548 & 5,206 & 1,554 & 0.430\\
			Cold & 1,084 & 1,214 & 565 & 519 & 649 & 0.492\\
			Wind & 2,676 & 5,096 & 1,902 & 774 & 3,194 & 0.489\\
			\bottomrule
		\end{tabular}
	\end{table}
	
	Event support decreases sharply with depth. At $k=2$, only 62, 10 and 112 ERA5 heat, cold and wind exceedances remain, respectively, compared with 25, 23 and 467 under HRES-fc0. These are spatially and temporally dependent cell--time cases, not independent events. The deepest conditional curves, particularly for cold, therefore describe very limited samples.
	\FloatBarrier
	
	\subsection{All-case and record-conditioned errors at 48 hours}
\begin{figure}[htbp]
		\centering
		\includegraphics[width=\textwidth]{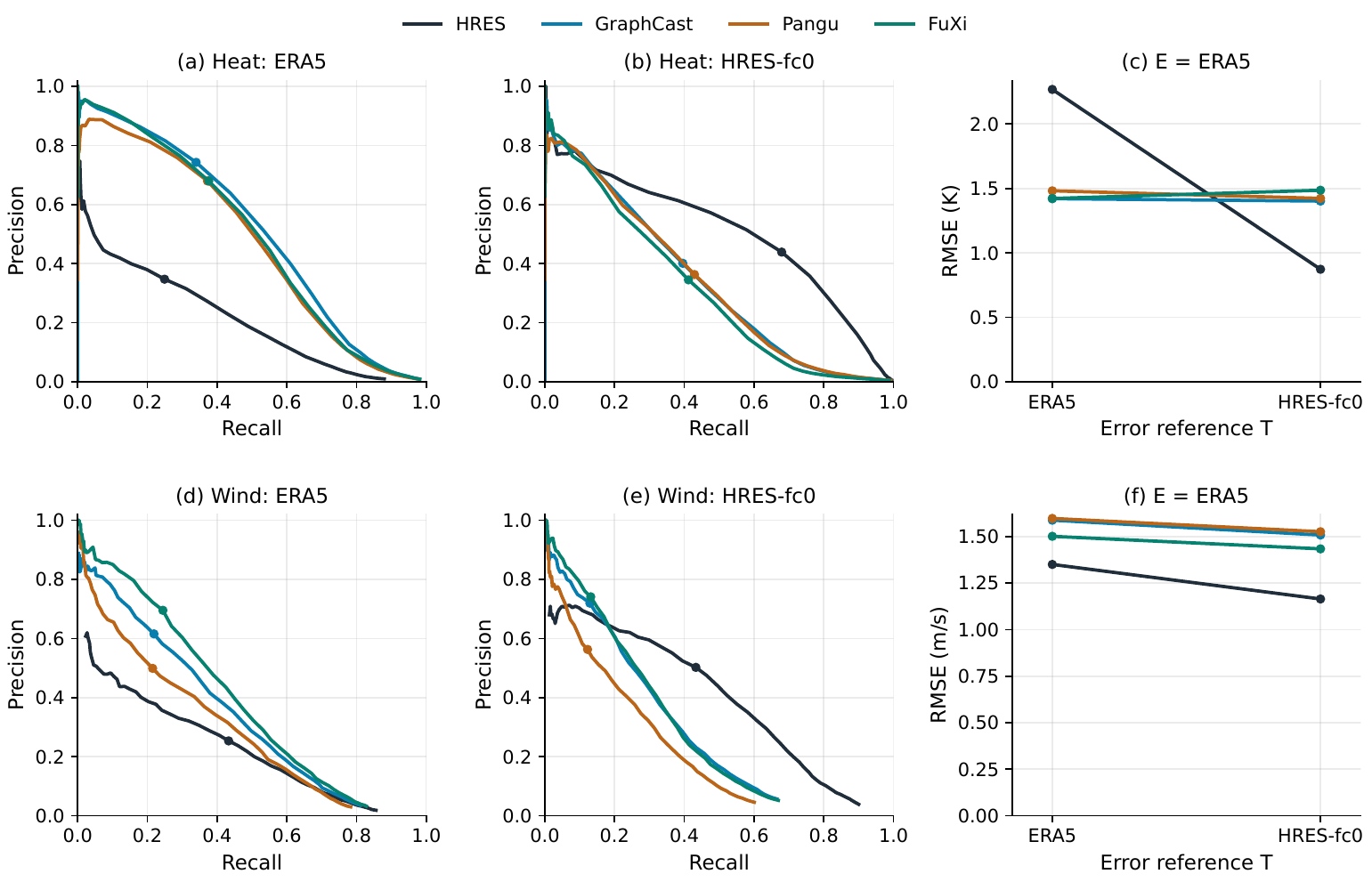}
		\caption{Temperature ranking sensitivity to the verification reference for 48-hour forecasts. Left and centre: area-weighted precision--recall curves under each event reference, obtained by shifting forecast detection thresholds; dots mark the issued, unshifted forecasts. Right: record-conditioned RMSE with event selection fixed to ERA5 and only the error reference changed. Temperature rankings reverse, while HRES retains its wind RMSE advantage. The threshold sweeps are descriptive controls, not fitted postprocessing.}
		\label{fig:reference2020}
	\end{figure}

	Figure~\ref{fig:reference2020} summarizes the 48-hour occurrence and intensity comparisons. Table~\ref{tab:detRMSE} compares all-case RMSE with record-conditioned RMSE under the four reference combinations. For temperature, the best-performing model changes when only the error reference changes, under either event selector. This sensitivity is already present before selecting records. On ERA5-selected heat and cold records, every AI model outperforms HRES against ERA5, while HRES outperforms every AI model against HRES-fc0.
	
	Wind behaves differently on the observed-record subsets. Although all three AI models improve all-case wind RMSE against ERA5, HRES has the lowest record-conditioned RMSE under all four reference combinations. The corresponding record-boundary comparison also favours HRES at 120 and 240~h in the numerical tables. This result concerns errors on observed exceedances and does not imply superiority under every occurrence metric or selection rule.
	
	\begin{table}[htbp]
		\centering\small
		\caption{Area-weighted RMSE at 48~h in 2020, on all eligible cases and on observed record exceedances ($k=0$). E denotes ERA5 and H denotes HRES-fc0. All-case columns specify the error reference $T$; record columns specify $(E,T)$, where $E$ selects events and $T$ measures errors. Every column compares models on a common sample against a common target. Heat and cold use kelvin; wind uses m\,s$^{-1}$. The all-case temperature values are repeated in the heat and cold blocks because they use the same full temperature sample. Bold indicates the lowest RMSE within each extreme and column.}
		\label{tab:detRMSE}\label{tab:detAllRMSE}
		\begin{tabular}{llrrrrrr}
			\toprule
			& & \multicolumn{2}{c}{All cases} & \multicolumn{4}{c}{Observed records}\\
			\cmidrule(lr){3-4}\cmidrule(lr){5-8}
			Extreme & Model & $T=\mathrm{E}$ & $T=\mathrm{H}$ & $(E,E)$ & $(E,H)$ & $(H,E)$ & $(H,H)$\\
			\midrule
			Heat & HRES & 1.602 & \textbf{0.956} & 2.268 & \textbf{0.873} & 1.662 & \textbf{1.156}\\
			& GraphCast & \textbf{0.975} & 1.355 & \textbf{1.420} & 1.402 & \textbf{1.269} & 1.803\\
			& Pangu & 1.083 & 1.333 & 1.481 & 1.423 & 1.318 & 1.846\\
			& FuXi & 1.016 & 1.407 & 1.422 & 1.486 & 1.325 & 1.895\\
			\midrule
			Cold & HRES & 1.602 & \textbf{0.956} & 2.508 & \textbf{1.250} & 1.321 & \textbf{1.375}\\
			& GraphCast & \textbf{0.975} & 1.355 & 1.995 & 1.685 & 1.212 & 1.768\\
			& Pangu & 1.083 & 1.333 & 2.257 & 1.703 & 1.387 & 1.857\\
			& FuXi & 1.016 & 1.407 & \textbf{1.907} & 1.693 & \textbf{1.155} & 1.680\\
			\midrule
			Wind & HRES & 0.647 & \textbf{0.589} & \textbf{1.350} & \textbf{1.164} & \textbf{0.990} & \textbf{1.008}\\
			& GraphCast & 0.586 & 0.693 & 1.588 & 1.508 & 1.089 & 1.519\\
			& Pangu & 0.629 & 0.731 & 1.597 & 1.526 & 1.137 & 1.599\\
			& FuXi & \textbf{0.568} & 0.675 & 1.501 & 1.434 & 1.029 & 1.479\\
			\bottomrule
		\end{tabular}
	\end{table}
	\FloatBarrier
	
	\subsection{Record occurrence}
	The issued 48-hour temperature classifications favour the AI models in F1 when ERA5 defines events, and HRES when HRES-fc0 defines events (Table~\ref{tab:detOccurrence48}). Precision and recall explain how those rankings arise. Against ERA5, the AI cold forecasts are more precise but less sensitive than HRES, whereas their heat forecasts improve both measures. Under HRES-fc0, HRES has higher recall for both temperature extremes and the highest F1.
	
\begin{figure}[p]
		\centering\includegraphics[width=\textwidth,height=.83\textheight,keepaspectratio]{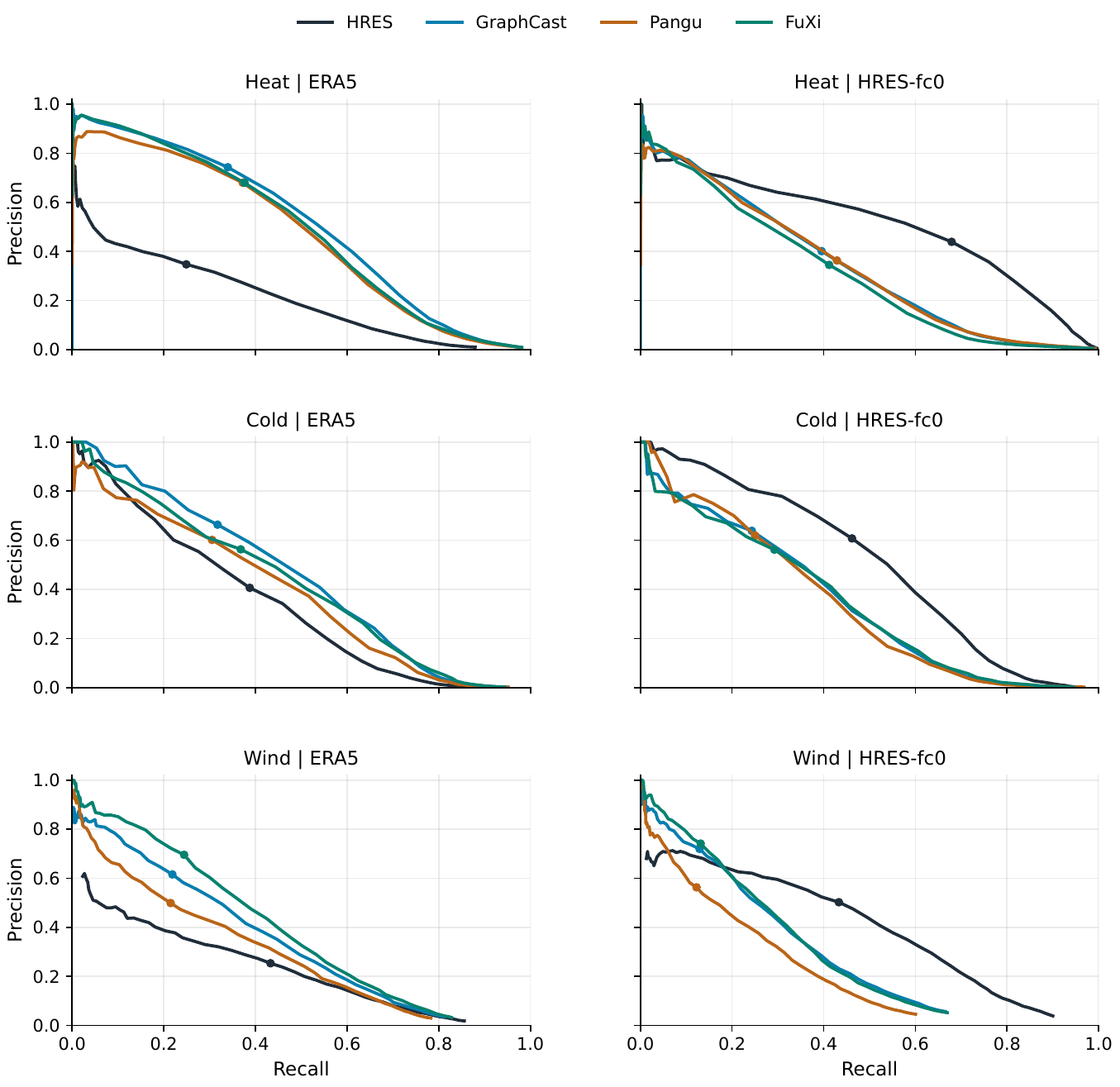}
		\caption{Area-weighted precision--recall curves at 48~h for heat, cold and wind under the two references. The finite forecast-shift sweep changes detection thresholds; filled dots show the issued, unshifted classifications summarized in Table~\ref{tab:detOccurrence48}. }
		\label{fig:detPR48}
	\end{figure}

	For wind against ERA5, the AI forecasts also trade lower recall for higher precision. FuXi achieves the highest F1 despite having larger record-conditioned intensity errors than HRES. Under HRES-fc0, HRES has the highest wind F1, reflecting its stronger recall. The precision--recall sweeps in Figure~\ref{fig:detPR48} extend these comparisons beyond the issued threshold without implying a single ranking at all decision thresholds.
	
	\begin{table}[htbp]
		\centering\small
		\caption{Area-weighted precision, recall and F1 at 48~h in 2020 for the issued, unshifted record forecasts. Reference events are crossings of the fixed ERA5 historical boundary. Bold identifies the largest F1 within each extreme and reference; precision and recall are shown separately to expose their trade-off.}
		\label{tab:detOccurrence48}
		\begin{tabular}{llrrrrrr}
			\toprule
			& & \multicolumn{3}{c}{ERA5 events} & \multicolumn{3}{c}{HRES-fc0 events}\\
			\cmidrule(lr){3-5}\cmidrule(lr){6-8}
			Extreme & Model & Precision & Recall & F1 & Precision & Recall & F1\\
			\midrule
			Heat & HRES & 0.347 & 0.249 & 0.290 & 0.439 & 0.679 & \textbf{0.533}\\
			& GraphCast & 0.743 & 0.340 & 0.466 & 0.401 & 0.396 & 0.398\\
			& Pangu & 0.680 & 0.372 & 0.481 & 0.363 & 0.429 & 0.393\\
			& FuXi & 0.680 & 0.376 & \textbf{0.484} & 0.345 & 0.412 & 0.376\\
			\midrule
			Cold & HRES & 0.406 & 0.388 & 0.397 & 0.608 & 0.462 & \textbf{0.525}\\
			& GraphCast & 0.664 & 0.317 & 0.429 & 0.639 & 0.243 & 0.352\\
			& Pangu & 0.602 & 0.305 & 0.405 & 0.619 & 0.250 & 0.356\\
			& FuXi & 0.563 & 0.368 & \textbf{0.445} & 0.562 & 0.292 & 0.385\\
			\midrule
			Wind & HRES & 0.253 & 0.433 & 0.320 & 0.502 & 0.433 & \textbf{0.465}\\
			& GraphCast & 0.616 & 0.219 & 0.323 & 0.720 & 0.129 & 0.219\\
			& Pangu & 0.499 & 0.215 & 0.300 & 0.563 & 0.122 & 0.201\\
			& FuXi & 0.695 & 0.244 & \textbf{0.362} & 0.741 & 0.132 & 0.223\\
			\bottomrule
		\end{tabular}
	\end{table}
	
	Forecast crossing counts provide complementary context (Table~\ref{tab:detCounts}). The AI systems all forecast fewer heat, cold and wind crossings than HRES at 48~h. At longer leads, the most pronounced contraction occurs for FuXi's cold and wind forecasts, which produce very few crossings at 240~h. The pattern is model dependent: HRES and Pangu wind counts instead increase with lead. Counts alone cannot distinguish successful detections from false alarms or establish which system has better occurrence skill.
	
	\begin{table}[htbp]
		\centering\small
		\caption{Forecast record-crossing counts (unweighted) in 2020 on the same 4,649,708 land cell--time cases at every lead. Counts depend on the forecast and the fixed ERA5 historical boundary, not on the subsequent choice of verifying reference. Reference crossing counts are given in Table~\ref{tab:overlap2020}.}
		\label{tab:detCounts}
		\begin{tabular}{lrrrrrrrrr}
			\toprule
			& \multicolumn{3}{c}{Heat} & \multicolumn{3}{c}{Cold} & \multicolumn{3}{c}{Wind}\\
			\cmidrule(lr){2-4}\cmidrule(lr){5-7}\cmidrule(lr){8-10}
			Model & 48~h & 120~h & 240~h & 48~h & 120~h & 240~h & 48~h & 120~h & 240~h\\
			\midrule
			HRES & 5,489 & 5,219 & 4,874 & 1,066 & 1,189 & 914 & 4,438 & 4,533 & 4,681\\
			GraphCast & 3,888 & 2,604 & 1,966 & 562 & 625 & 1,159 & 1,210 & 810 & 681\\
			Pangu & 4,728 & 3,732 & 4,100 & 561 & 556 & 460 & 1,344 & 1,369 & 1,452\\
			FuXi & 4,616 & 4,143 & 1,158 & 694 & 705 & 19 & 1,096 & 795 & 38\\
			\bottomrule
		\end{tabular}
	\end{table}
	\FloatBarrier
	
	\clearpage
	
	\subsection{Joint forecast record control}
	Table~\ref{tab:detJoint48} compares each AI model with HRES on cases where both forecasts cross the historical boundary. All three heat comparisons favour AI against ERA5 and HRES against HRES-fc0, so the heat reversal persists without selecting on the verifying outcome. Cold comparisons are more mixed against ERA5 and consistently favour HRES against HRES-fc0.
	
	Wind illustrates the importance of the selection rule. On these jointly forecast subsets, the AI models have lower RMSE than HRES against ERA5, while the HRES-fc0 comparison is mixed. This differs from HRES's uniform wind advantage on observed-record subsets in Table~\ref{tab:detRMSE}. The control selects different cases for each AI--HRES pair, however, and therefore supports within-pair comparisons rather than an overall ranking across AI models.
	
	\begin{table}[htbp]
		\centering\small
		\caption{Area-weighted RMSE at 48~h where the named AI model and HRES both forecast a record. $N$ is the (unweighted) number of selected cell--time cases. Each row uses the same subset under both error references, but subsets differ across rows. Heat and cold use kelvin; wind uses m\,s$^{-1}$. Bold identifies the lower RMSE within each row and reference.}
		\label{tab:detJoint48}
		\begin{tabular}{llrrrrr}
			\toprule
			& & & \multicolumn{2}{c}{ERA5 errors} & \multicolumn{2}{c}{HRES-fc0 errors}\\
			\cmidrule(lr){4-5}\cmidrule(lr){6-7}
			Extreme & AI model & $N$ & AI & HRES & AI & HRES\\
			\midrule
			Heat & GraphCast & 1,716 & \textbf{1.065} & 1.313 & 1.092 & \textbf{0.900}\\
			& Pangu & 1,885 & \textbf{1.082} & 1.355 & 1.116 & \textbf{0.926}\\
			& FuXi & 1,790 & \textbf{1.065} & 1.308 & 1.170 & \textbf{0.908}\\
			\midrule
			Cold & GraphCast & 370 & 1.079 & \textbf{1.078} & 1.105 & \textbf{0.711}\\
			& Pangu & 355 & 1.255 & \textbf{1.197} & 1.122 & \textbf{0.733}\\
			& FuXi & 404 & \textbf{1.081} & 1.211 & 1.005 & \textbf{0.726}\\
			\midrule
			Wind & GraphCast & 824 & \textbf{0.895} & 1.110 & 0.995 & \textbf{0.983}\\
			& Pangu & 709 & \textbf{0.905} & 1.030 & \textbf{0.938} & 0.952\\
			& FuXi & 730 & \textbf{0.763} & 0.942 & 0.932 & \textbf{0.855}\\
			\bottomrule
		\end{tabular}
	\end{table}
	\FloatBarrier
	
	\subsection{Conditional errors and outcome selection}
	\label{sec:detconditional}
	
\begin{figure}[p]
		\centering\includegraphics[width=\textwidth,height=.83\textheight,keepaspectratio]{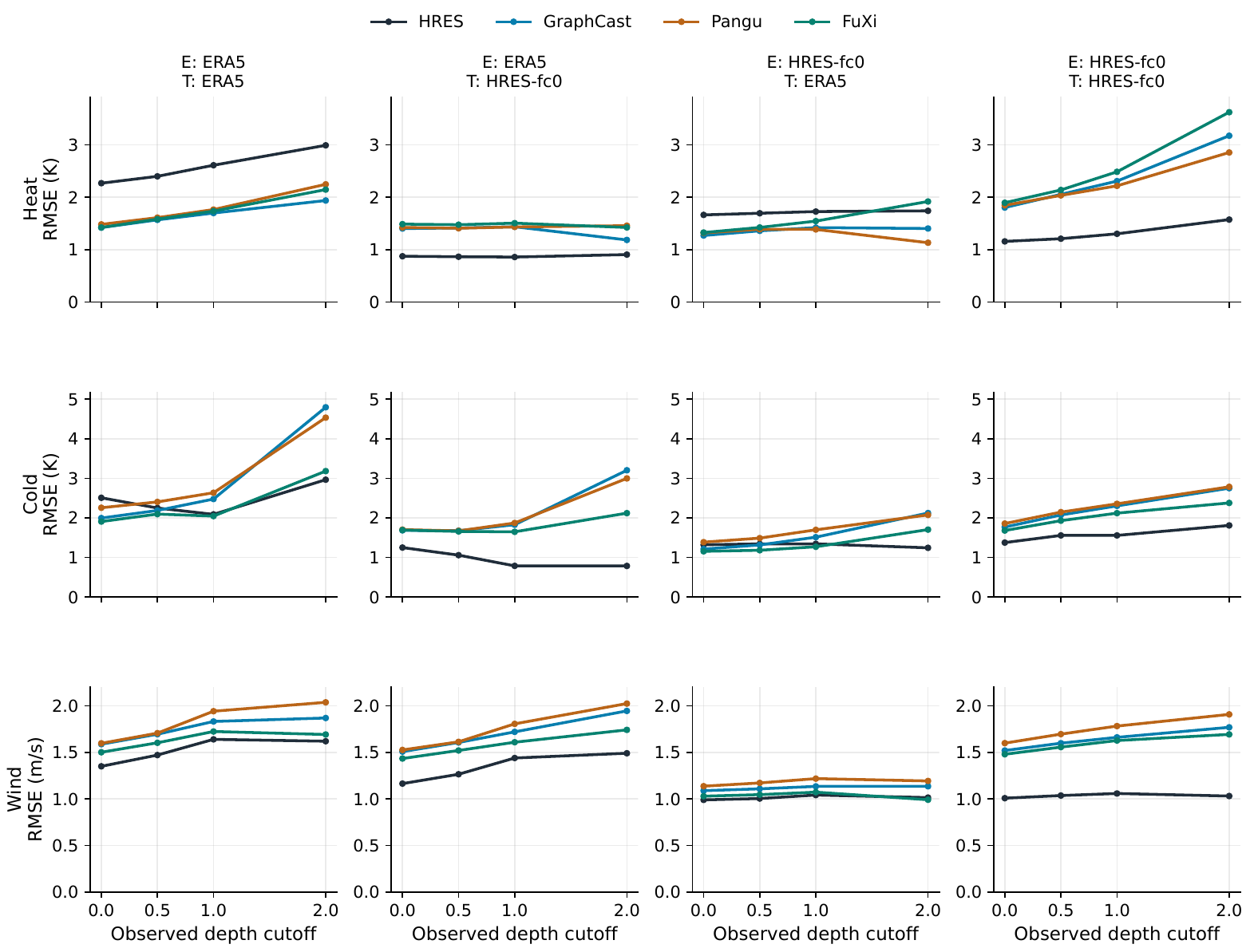}
		\caption{Conditional RMSE at 48~h as a function of standardized observed depth, for all four $(E,T)$ combinations. The scale is the historical standard deviation of annual monthly extrema. Each point uses cases exceeding the plotted cutoff; sparse deep subsets, especially cold, require caution.}
		\label{fig:detdepthrmse48}
	\end{figure}

\begin{figure}[p]
		\centering\includegraphics[width=\textwidth,height=.83\textheight,keepaspectratio]{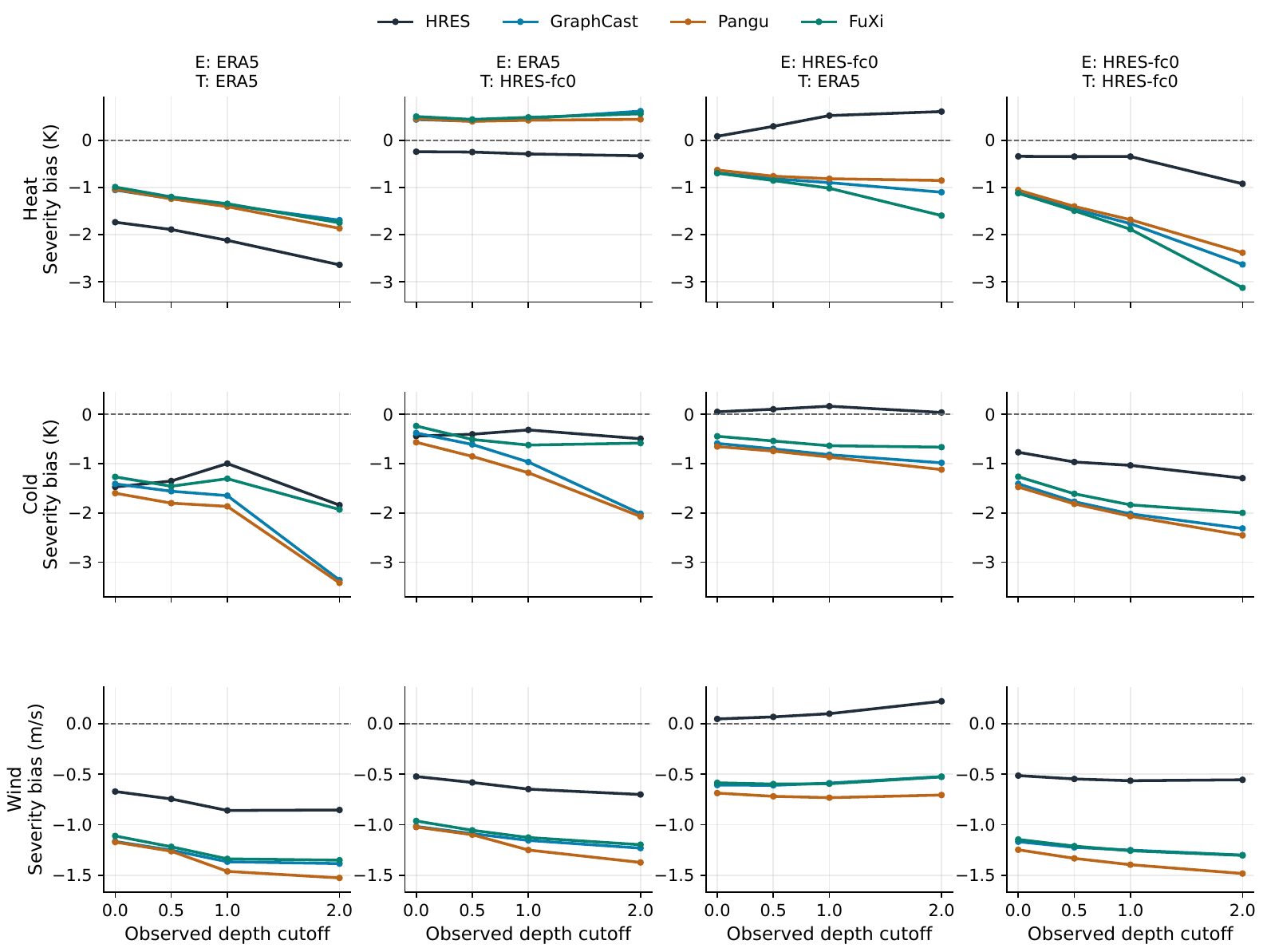}
		\caption{Conditional severity bias at 48~h as a function of standardized observed depth, for all four $(E,T)$ combinations. Cold errors are reflected, so negative bias denotes insufficient extremeness for every variable. Selection and scaling follow Figure~\ref{fig:detdepthrmse48}.}
		\label{fig:detdepthbias48}
	\end{figure}

\begin{figure}[p]
		\centering\includegraphics[width=\textwidth,height=.83\textheight,keepaspectratio]{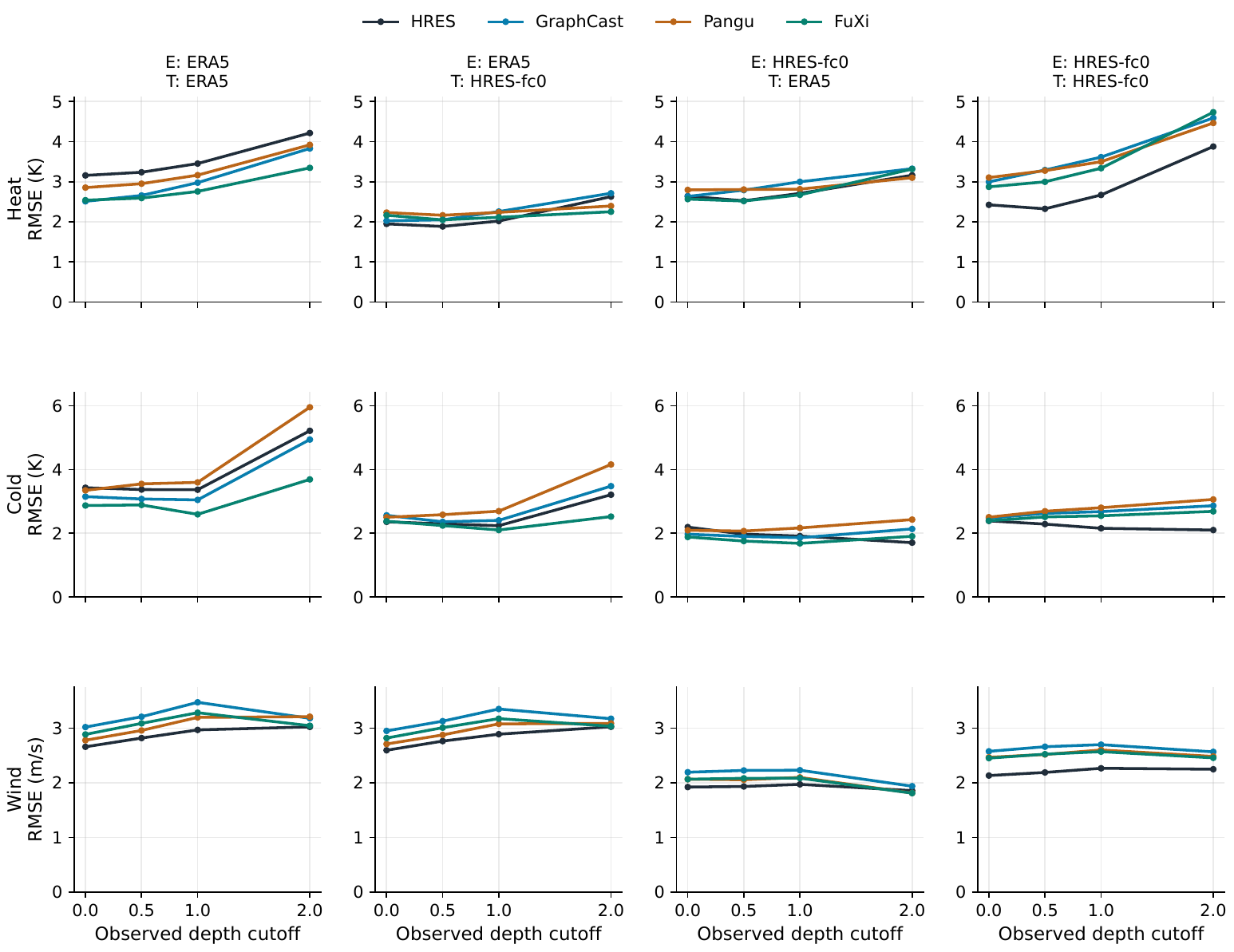}
		\caption{Conditional RMSE at 120~h as a function of standardized observed depth. Event selection, error references and historical scales follow Figure~\ref{fig:detdepthrmse48}, providing a longer-lead comparison on the same eligible sample.}
		\label{fig:detdepthrmse120}
	\end{figure}

\begin{figure}[p]
		\centering\includegraphics[width=\textwidth,height=.83\textheight,keepaspectratio]{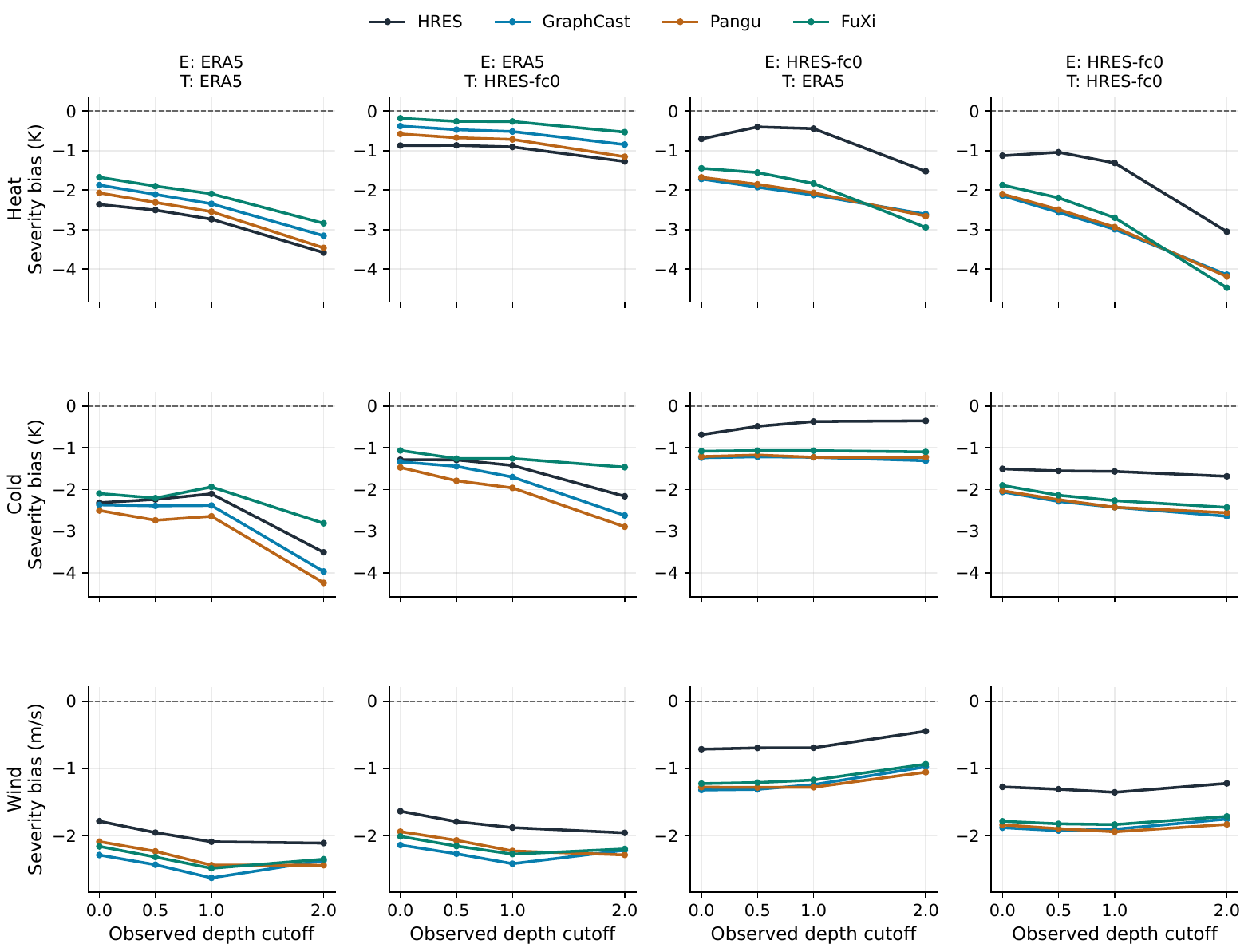}
		\caption{Conditional severity bias at 120~h as a function of standardized observed depth. The four reference combinations, cumulative selection rule and cold sign convention follow Figure~\ref{fig:detdepthbias48}.}
		\label{fig:detdepthbias120}
	\end{figure}

	Figures~\ref{fig:detdepthrmse48} and~\ref{fig:detdepthbias48}
	show conditional RMSE and severity bias at 48~h;
	Figures~\ref{fig:detdepthrmse120} and~\ref{fig:detdepthbias120}
	show the corresponding 120-hour results. Each point shows the cases
	beyond the standardized-depth cutoff. Increasing depth
	therefore changes both the size and composition of the evaluated
	sample.
	
	Selecting observed extremes can produce systematic underprediction,
	even for an ideal conditional-mean forecast. Let $Z_i$ denote the
	verifying variable, oriented so that larger values indicate greater
	severity, with cold reflected as in the Methods. Let $\mathcal I_i$
	contain the information available for the forecaster. The ideal mean
	forecast is $\mu_i=\E[Z_i\mid\mathcal I_i]$, and its expected error
	before selecting outcomes is
	\[
	\E[\mu_i-Z_i\mid\mathcal I_i]=0.
	\]
	This expectation averages over all possible outcomes given the
	forecast information, including both unusually high and unusually
	low realizations.
	
	Now retain only outcomes exceeding a known threshold
	$u_i(k)=r_i+k s_i$, with $s_i>0$. The forecast $\mu_i$ averages over
	both the higher outcomes retained by this selection and the lower
	outcomes discarded by it. The retained group consequently has a
	higher mean than $\mu_i$, provided that both groups have positive
	conditional probability and the conditional first moments are
	finite. The expected forecast error on the retained group is
	therefore negative:
	\begin{equation}
		b_i(k)
		:=\E[\mu_i-Z_i\mid\mathcal I_i,Z_i>u_i(k)]
		=\mu_i-\E[Z_i\mid\mathcal I_i,Z_i>u_i(k)]<0.
		\label{eq:idealbias}
	\end{equation}
	The forecast remains unchanged; selection has increased the mean
	of the outcomes against which it is evaluated. This argument
	does not require the threshold itself to exceed $\mu_i$.
	
	Raising the threshold removes the lowest outcomes from the
	already selected group, so its conditional mean cannot decrease.
	At fixed forecast information, $\mu_i$ stays constant and the
	conditional bias therefore becomes more negative or remains
	unchanged. For $k_2>k_1$ satisfying the conditions above,
	\begin{equation}
		b_i(k_2)\le b_i(k_1)<0.
		\label{eq:idealbiasdepth}
	\end{equation}
	Equality is possible when the higher cutoff removes no probability
	mass. Increasing underprediction can thus arise through outcome
	selection alone, without a structural cap on forecast values.
	This illustrates the selection effect underlying the forecaster's
	dilemma \citep{lerch2017forecasters}.
	
	These statements hold at fixed forecast information and when
	selection and errors concern the same verifying target. Empirical
	depth curves pool different locations, dates and forecast situations,
	whose relative contributions change with depth. They therefore
	need not be monotone, and the argument does not prescribe the sign
	or shape of crossed-reference comparisons where $E\ne T$.
	Conditional RMSE additionally depends on error variability.
	The curves remain useful descriptions of performance on realized
	extremes, but increasingly negative bias alone does not establish
	structural extrapolation failure.
	

    \newpage
	
    \section{Supplementary tables}
	
    \begin{table}[htbp]
		\centering\small
		\caption{Observed 2022 land grid-cell--time exceedances. Counts are unweighted and identical across models and leads on the common sample. }
		\label{tab:counts}
		\begin{tabular}{llrrrrrr}
			\toprule
			Reference & Extreme & $k=0$ & $0.5$ & $1$ & $2$ & $2.5$ & $3$\\
			\midrule
			ERA5 & Heat & 5,129 & 1,772 & 610 & 28 & 4 & ---\\
			& Cold & 1,665 & 764 & 446 & 203 & 157 & 114\\
			& Wind & 2,971 & 1,437 & 688 & 200 & 115 & 59\\
			\midrule
			HRES-fc0 & Heat & 3,514 & 1,369 & 509 & 21 & 4 & 1\\
			& Cold & 2,275 & 1,013 & 561 & 261 & 183 & 130\\
			& Wind & 5,838 & 3,153 & 1,727 & 552 & 324 & 206\\
			\bottomrule
		\end{tabular}
	\end{table}

    \newpage
    
	\section{Supplementary figures}
	
	\begin{figure}[p]
		\centering\includegraphics[width=\textwidth,height=.83\textheight,keepaspectratio]{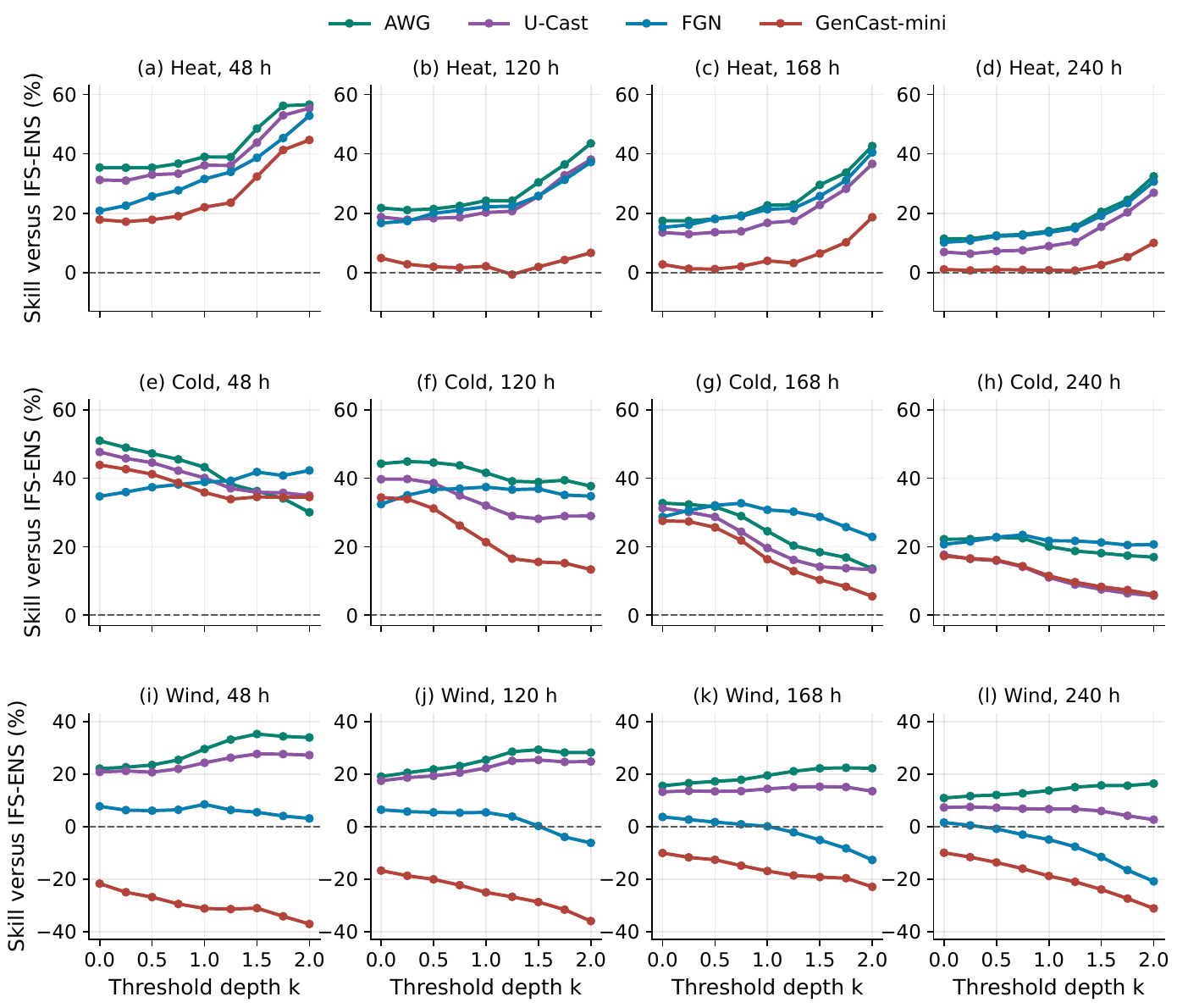}
		\caption{Fair record Brier skill against ERA5. All cases contribute at every threshold. Heat remains favourable to all AI systems, while some deep-wind comparisons lose skill.}
		\label{fig:bsERA}
	\end{figure}
	\begin{figure}[p]
		\centering\includegraphics[width=\textwidth,height=.83\textheight,keepaspectratio]{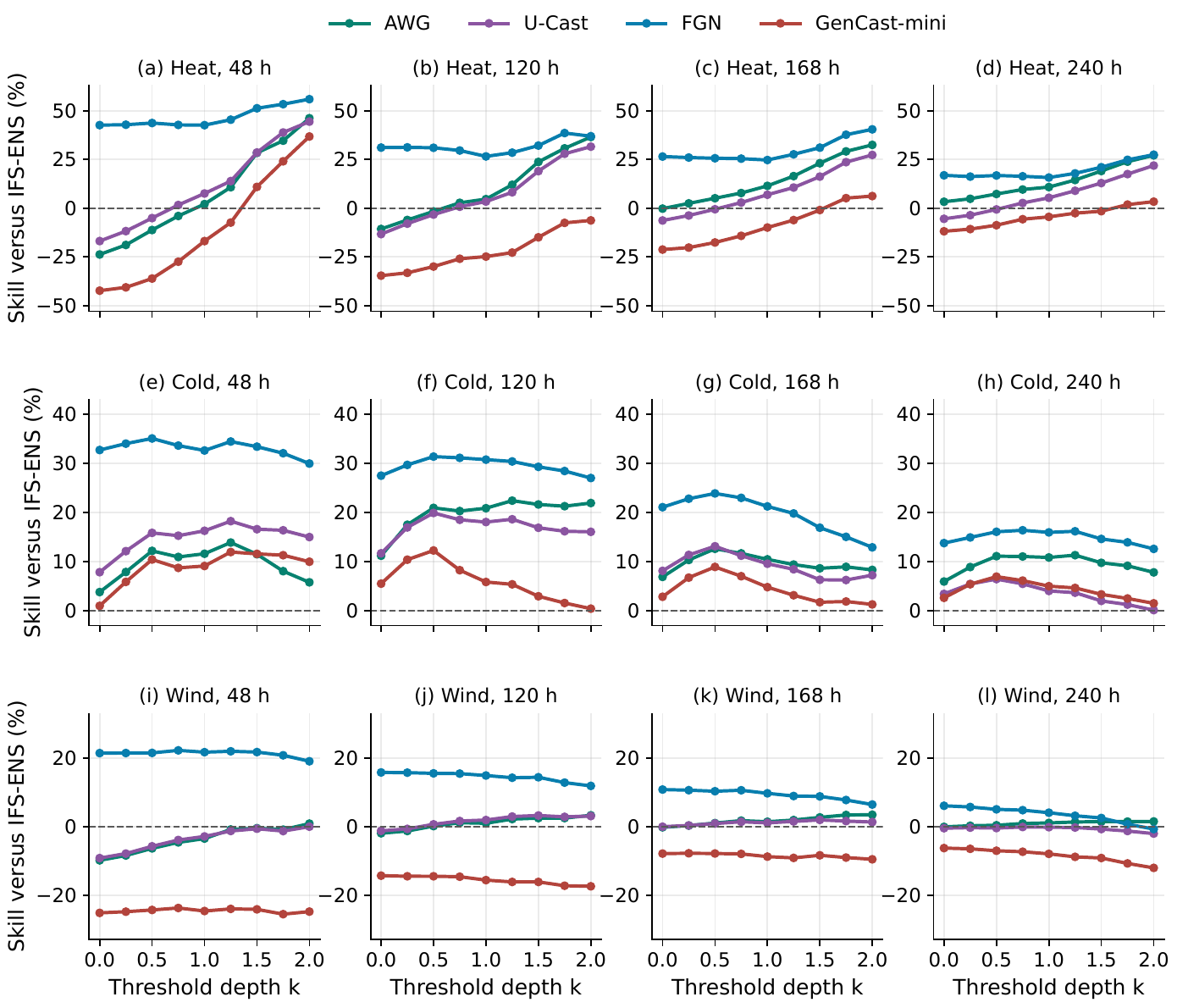}
		\caption{Fair record Brier skill against HRES-fc0 using ERA5 historical thresholds. Negative near-record heat skill for AWG and U-Cast becomes positive at greater depth; wind is more mixed.}
		\label{fig:bsHRES}
	\end{figure}
	\begin{figure}[p]
		\centering\includegraphics[width=\textwidth,height=.83\textheight,keepaspectratio]{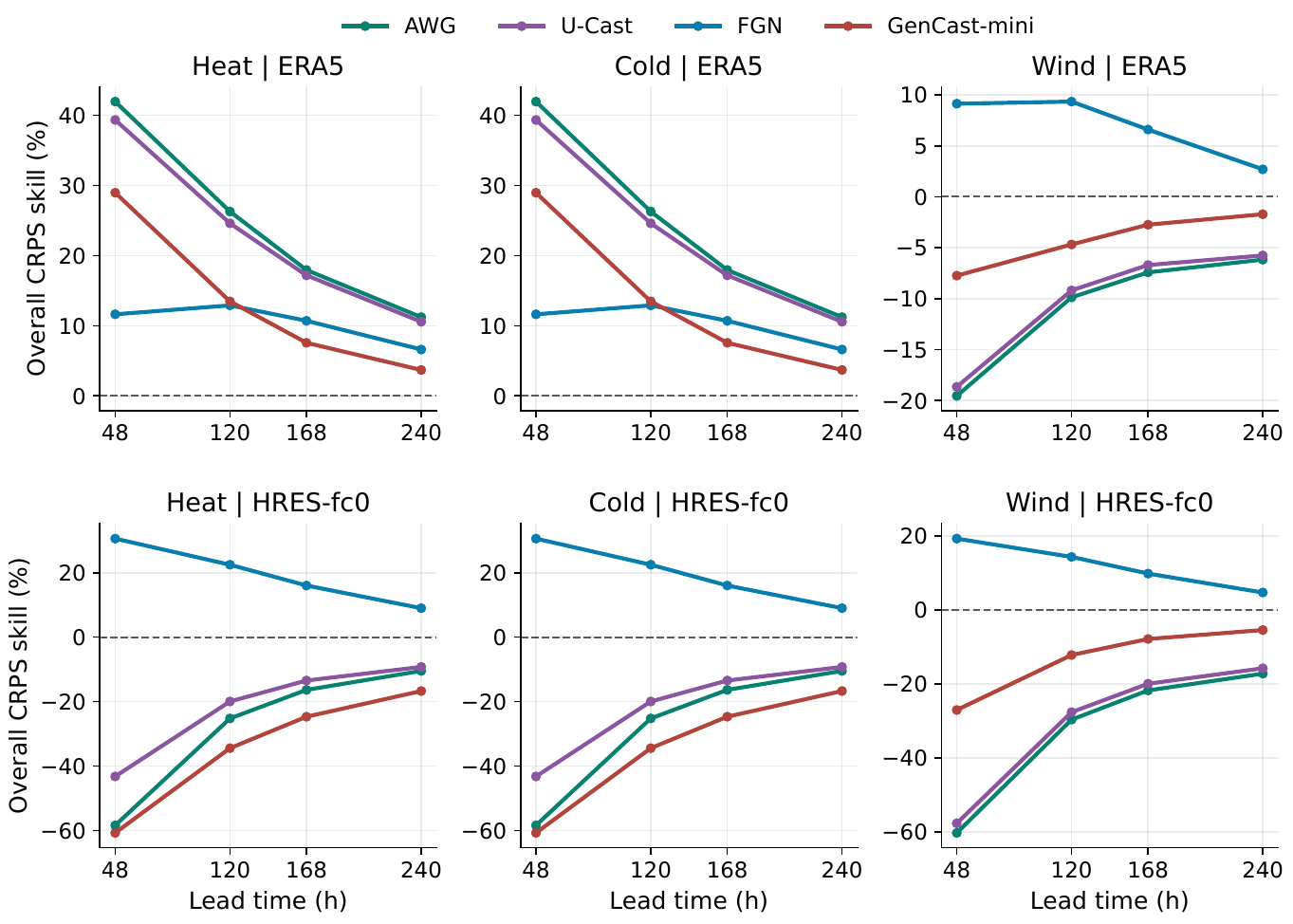}
		\caption{Ordinary fair CRPS skill on all cases under both common references. Heat and cold panels coincide because they use the same untransformed temperature forecasts and verifying values.}
		\label{fig:overall}
	\end{figure}
	\begin{figure}[p]
		\centering\includegraphics[width=\textwidth,height=.83\textheight,keepaspectratio]{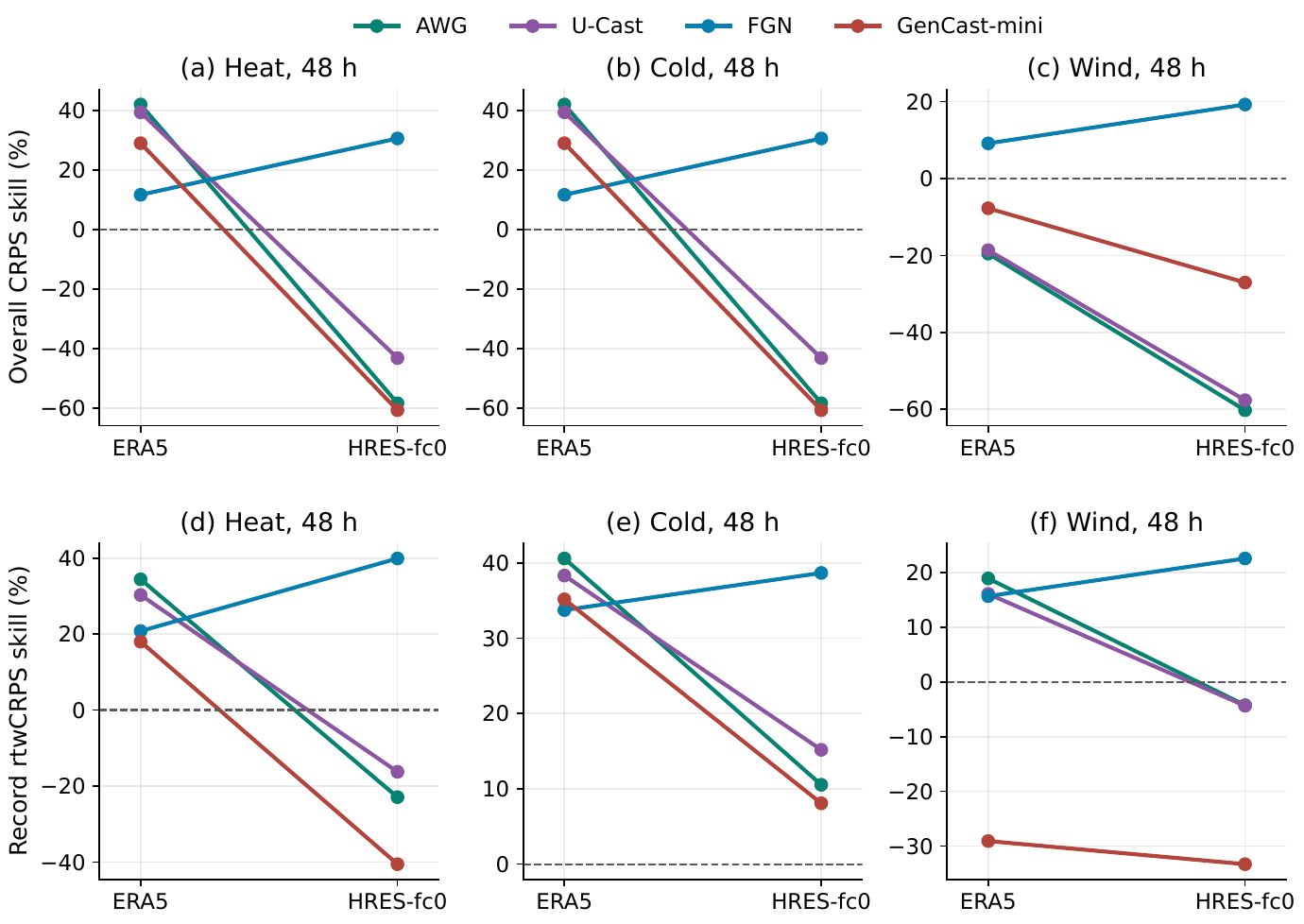}
		\caption{Reference switch at 48~h: ordinary fair CRPS (top) and fair rtwCRPS at $k=0$ (bottom). Each line connects one model's skill against the two references; each skill uses IFS-ENS verified against that same reference. }
		\label{fig:truthswitch}
	\end{figure}
	\begin{figure}[p]
		\centering\includegraphics[width=\textwidth,height=.83\textheight,keepaspectratio]{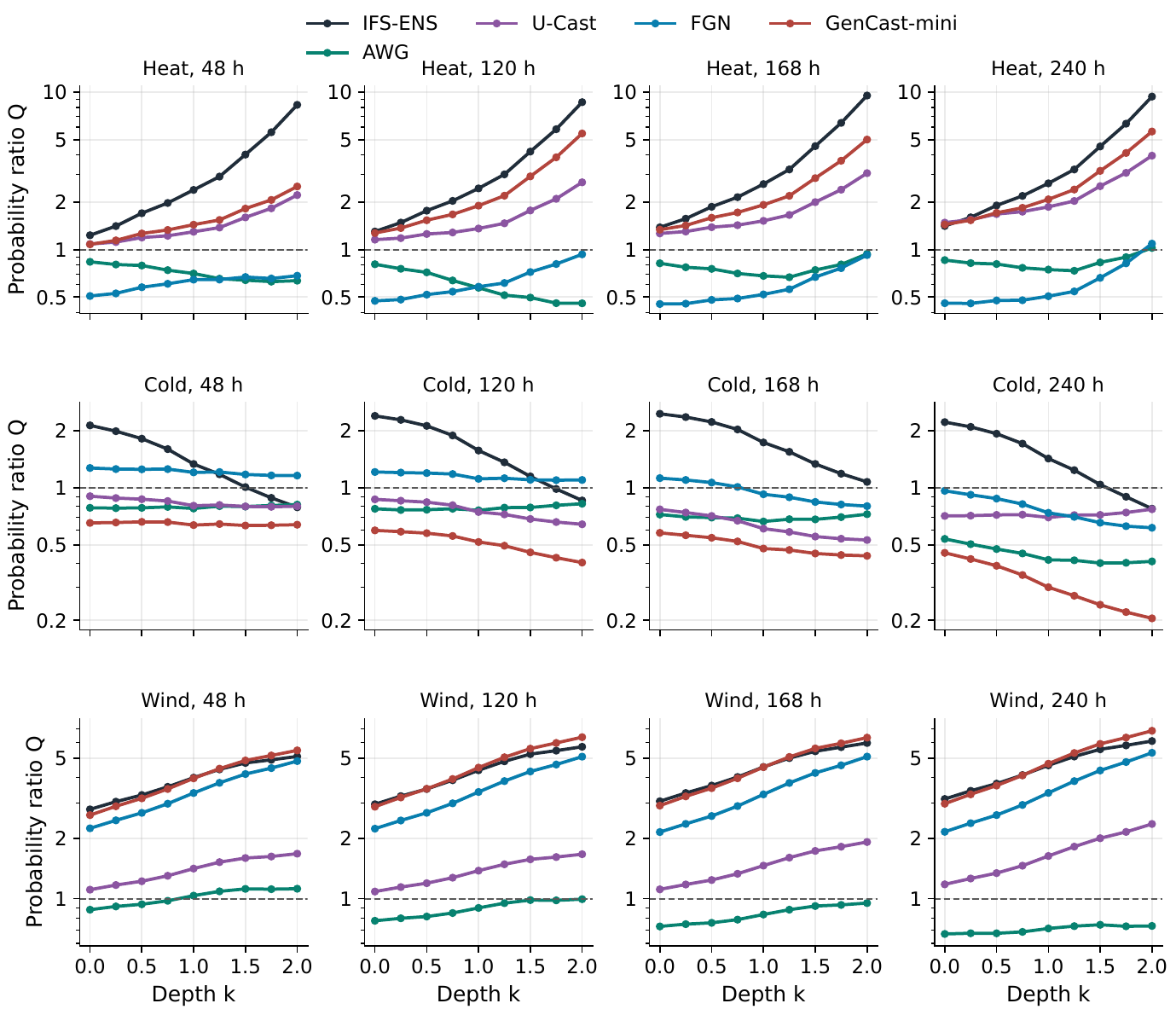}
		\caption{Aggregate forecast-probability ratios $Q$ against ERA5. The logarithmic vertical axis distinguishes probability deficits and surpluses relative to one.}
		\label{fig:qERA}
	\end{figure}
	\begin{figure}[p]
		\centering\includegraphics[width=\textwidth,height=.83\textheight,keepaspectratio]{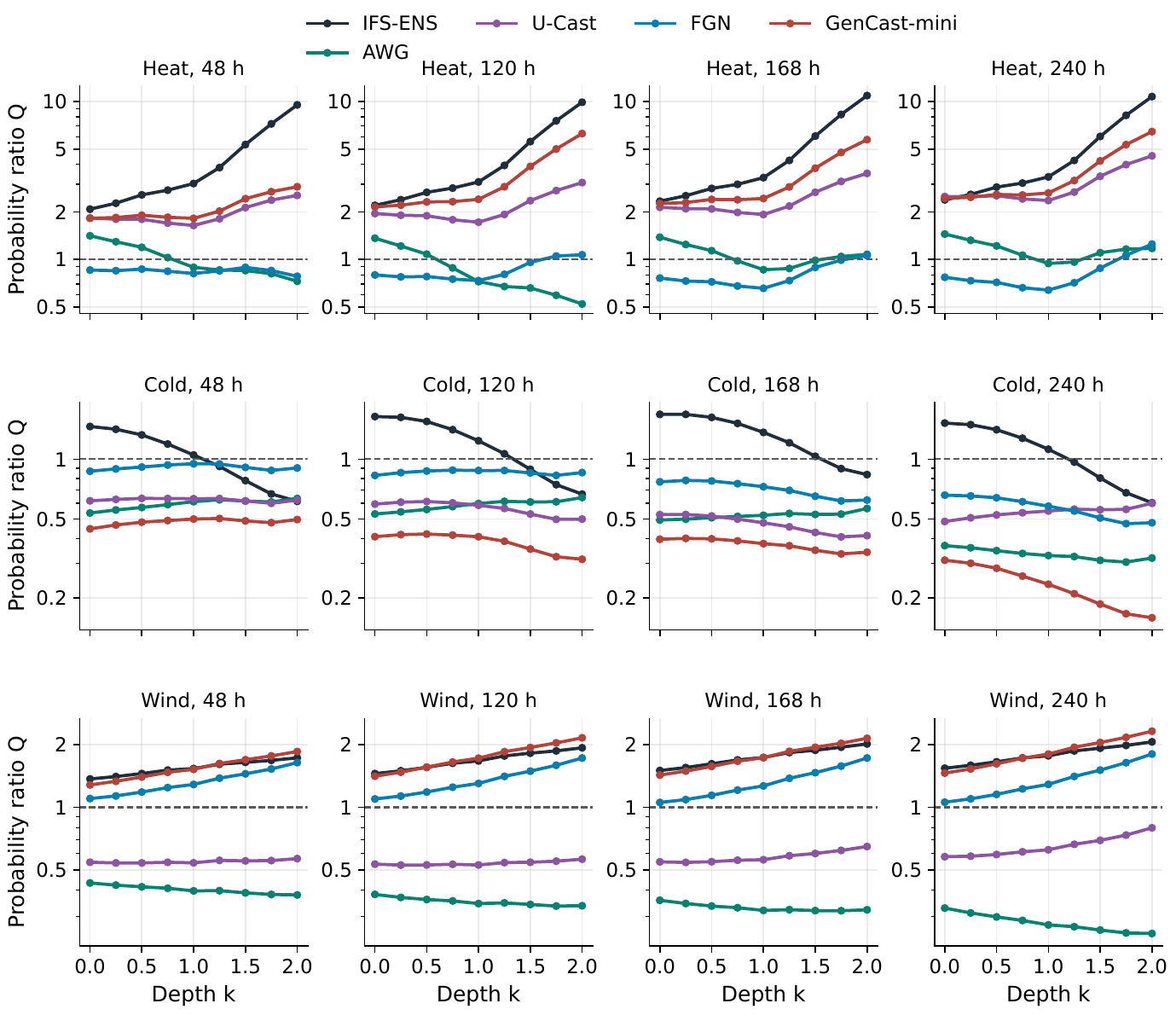}
		\caption{The same occurrence-frequency diagnostic against HRES-fc0. Thresholds and scales remain ERA5-based, while the verifying values and observed rates change.}
		\label{fig:qHRES}
	\end{figure}
	\begin{figure}[p]
		\centering\includegraphics[width=\textwidth,height=.83\textheight,keepaspectratio]{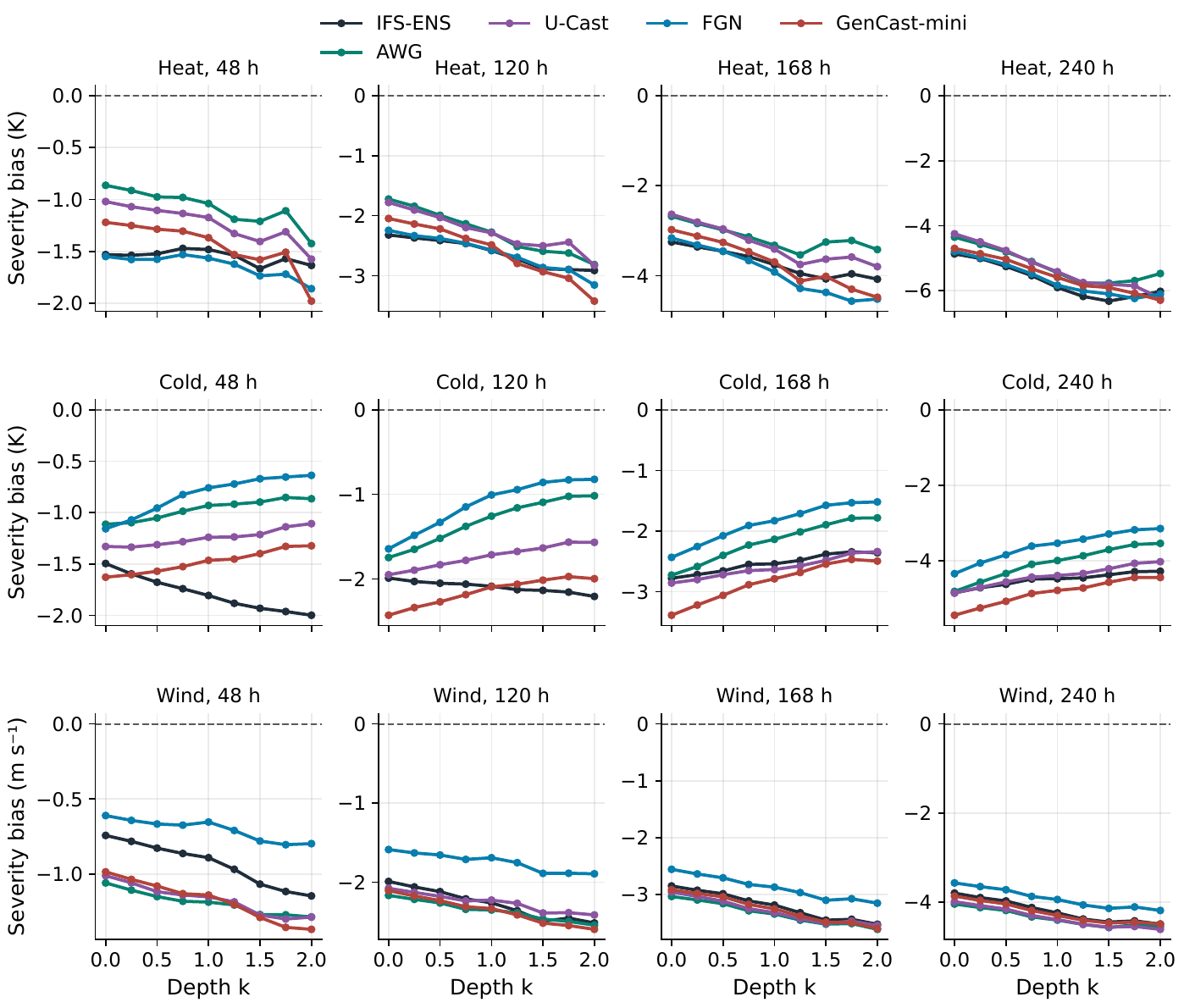}
		\caption{Ensemble-mean severity bias on realized exceedances against ERA5. Cold temperature bias is multiplied by $-1$, so negative values denote underprediction of extreme severity for all variables. }
		\label{fig:biasERA}
	\end{figure}
	\begin{figure}[p]
		\centering\includegraphics[width=\textwidth,height=.83\textheight,keepaspectratio]{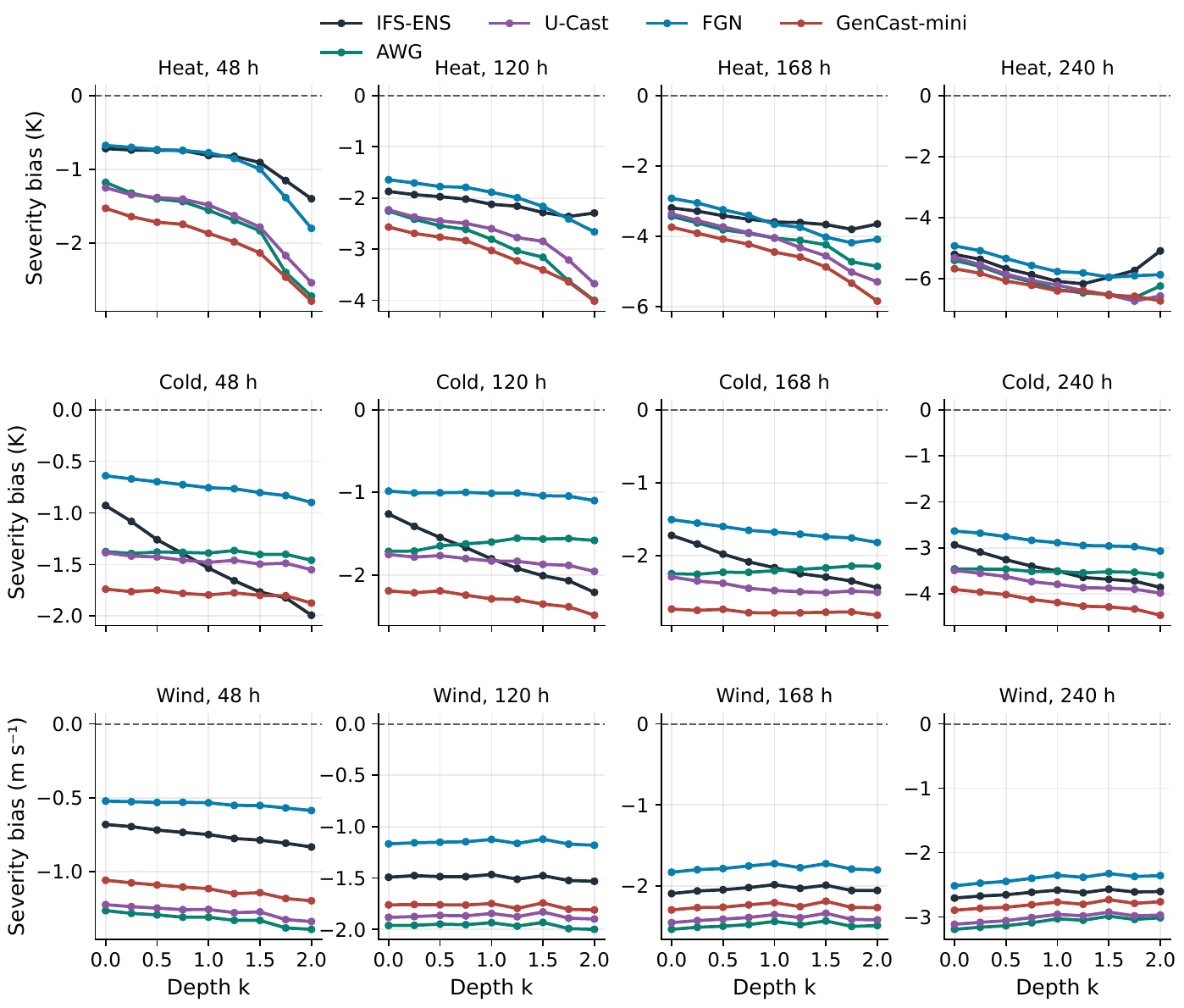}
		\caption{Conditional severity bias against HRES-fc0 with the same cold sign convention. Longer leads are shown for completeness; both predictive uncertainty and outcome selection affect these diagnostics.}
		\label{fig:biasHRES}
	\end{figure}

	\clearpage
\end{document}